\documentclass[twocolumn,twoside]{IEEEtran}
\IEEEoverridecommandlockouts 
\usepackage{graphicx} % for pdf, bitmapped graphics files
\usepackage{amsmath}
\usepackage{amssymb}
\usepackage{caption}
\usepackage{subcaption}
\usepackage{multirow}
\usepackage{multicol}
\usepackage{hyperref}
\newcommand*{\email}[1]{\href{mailto:#1}{\nolinkurl{#1}} } 
\usepackage[table]{xcolor}
\usepackage{array,multirow,graphicx}

\graphicspath{ {figs/} }
\usepackage{tikz}
\usepackage[normalem]{ulem}

\usepackage{lipsum}

\usepackage{tikz} 
\usetikzlibrary{matrix}

\usepackage{amsmath}
\usepackage[firstpageonly=true]{draftwatermark}

\begin{document}

\SetWatermarkAngle{0}
\SetWatermarkColor{black}
\SetWatermarkLightness{0.5}
\SetWatermarkFontSize{9pt}
% \SetWatermarkScale
% \SetWatermarkHorCenter
\SetWatermarkVerCenter{30pt}
\SetWatermarkText{\parbox{30cm}{%
\centering This is the authors' final version of the manuscript published as:\\
\centering M. Pliska et al., ``Single-Grasp Deformable Object Discrimination: The Effect of Gripper Morphology, Sensing Modalities, and Action Parameters,'' \\
\centering in IEEE Transactions on Robotics, vol. 40, pp. 4414-4426, 2024, (C) IEEE, \url{https://doi.org/10.1109/TRO.2024.3463402}. \\
\centering An accompanying video is available at \url{https://youtu.be/-6cmQrSzbCs?si=3-mxzMWyv8ZI6pMr}.
}}


\title{Single-grasp Deformable Object Discrimination: the Effect of Gripper Morphology, Sensing Modalities, and Action Parameters}

%\author[1]{Matej Hoffmann\texorpdfstring{\corref{cor1}}{}}\ead{matej.hoffmann@fel.cvut.cz} 

 \author{Michal~Pliska$^{1}$, Shubhan~Patni$^{1}$, Michal~Mares$^{1}$, Pavel~Stoudek$^{1}$,      Zdenek~Straka$^{1}$, Karla~Stepanova$^{2}$, and~Matej~Hoffmann$^{1}$   % <-this % stops a space
 \thanks{$^1$ Department of Cybernetics, Faculty of Electrical Engineering, Czech Technical University in Prague; \email{matej.hoffmann@fel.cvut.cz}.}
  \thanks{$^2$Czech Institute of Informatics, Robotics, and Cybernetics, Czech Technical University in Prague; \email{karla.stepanova@cvut.cz}.}
 \thanks{This work was co-funded by the European Union under the ROBOPROX (Robotics and advanced industrial production) project (reg. no. CZ.02.01.01/00/22\_008/0004590). K.S. was additionally supported by the Czech Science Foundation (project no.~GA21-31000S). The work originated in the IPALM project (H2020-FET-ERA-NET Cofund-CHIST-ERA III / TAČR EPSILON, no. TH05020001).}}% <-this % stops a space

\maketitle

% TRO - not more than 1200 chars
\begin{abstract}
In haptic object discrimination, the effect of gripper embodiment, action parameters, and sensory channels has not been systematically studied. We used two anthropomorphic hands and two 2-finger grippers to grasp two sets of deformable objects. On the object classification task, we found: (i) among classifiers, SVM on sensory features and LSTM on raw time series performed best across all grippers; (ii) faster compression speeds degraded performance; (iii) generalization to different grasping configurations was limited; transfer to different compression speeds worked well for the Barrett Hand only. Visualization of the feature spaces using PCA showed that gripper morphology and action parameters were the main source of variance, making generalization across embodiment or grip configurations very difficult. On the highly challenging dataset consisting of polyurethane foams alone, only the Barrett Hand achieved excellent performance. Tactile sensors can thus provide a key advantage even if recognition is based on stiffness rather than shape. The data set with 24,000 measurements is publicly available.
\end{abstract}


\section{Introduction}
\label{sec:intro}
%\deleted{deleted text}
%\added{added text}
%\replaced{new text}{old text}

Although visual object recognition has seen tremendous progress in the last years, not all object properties can be perceived using distal sensing. Physical object properties like stiffness, roughness, or mass are better acquired through manipulation.
Object recognition from haptic exploration can also be more robust, as it is insensitive to lighting conditions and object attributes that may be irrelevant for a task and may pose a challenge for visual perception (e.g., transparent objects \cite{li2023visual}). 
Deformable objects constitute an important case in which haptic recognition can be particularly effective. 

Sanchez et al.~\cite{sanchez2018robotic} provide a survey of robotic manipulation and sensing of deformable objects. Objects are considered deformable if they have (1) no compression strength (ropes and clothes) or (2) have a large strain\footnote{Young's modulus of elasticity smaller than around 10 MPa.} or have a large displacement. Additionally, a classification to 4 types based on geometry is presented. In this work, we leave objects of Type I-III (linear, planar, cloth-like) aside and focus on Type IV: triparametric objects---solid objects such as sponges or plush toys, which are also the least researched object type \cite{sanchez2018robotic}. 

Unlike works that employ robot hands with tactile arrays and complex haptic exploratory procedures (e.g., \cite{xu2013tactile,chu2013using}), we focus on single-grasp recognition \cite{spiers2016single} that should immediately classify the object---e.g., fruit based on its ripeness or plastic/paper/metal on a single-stream recycling line \cite{chin2019automated,guo2021visual,patni2024online}.
Such tasks are currently extremely labor intensive due to the need for manual object sorting. However, for example, plastic, paper, and metal differ significantly in their material properties and can be sorted through manipulation \cite{chin2019automated,patni2024online}.

Object recognition based on simulations of the interactions with deformable objects (see \cite{arriola2020modeling} for a review) has enormous computational costs that prevents online recognition.
The alternative are data-driven methods that directly process the time series generated by the interaction with the object. In this case, the detailed characteristics of the sensory streams become critical. This involves not only their number and type (force, touch, encoders/position sensors) but also the way they are generated---the gripper morphology and the parameters of the grasping action profoundly shape what is perceived (see \textit{morphology facilitating perception} in \cite{Mueller2017} for a theoretical account and \cite{Hoffmann_RAS_2014} for applying this idea to terrain recognition by a running robot). In this article, we study these effects on various setups and show that they are stronger than the choice of features or classifiers.

\textbf{Contribution.} First, we employed 4 different robot hands/grippers to grasp two sets of deformable objects. Our dataset comprising over 24000 measurements across different grippers, objects, and action parameters is made publicly available~\cite{osf2023single}. Second, depending on the gripper type, we analyze the effect of grasp parameters---finger configuration and compression speed---and the generalization of the trained classifiers to different action parameters. Third, we study the effect of individual sensory channels by performing object classification with a subset of the sensory channels. Fourth, we provide baseline discrimination performance using 4 different classifiers (feature-based $k$-NN and SVM; feature-less $k$-NN and LSTM) and demonstrate that most of the observations regarding the effect of gripper morphology, action parameters, and sensory channels generalize across the input type (raw time series or features) and classifier. Fifth, we perform an unsupervised analysis of the feature space (PCA) to assess the relative effects of the parameters under investigation (gripper, action parameters, and object).

\section{Related work}
\label{sec:rel_work}

Here we review the current works focused on recognition of deformable objects. We also review which sensory channels, features, classifiers, and exploratory actions were used in these works to classify objects based on deformation cues. 

\subsection{Stiffness-based object recognition}
Surveys of haptic or tactile robot perception are provided by \cite{luo2017robotic,li2020review}. The object’s \textit{hardness} might be probed by tapping on the object, while the \textit{object stiffness} or \textit{material elasticity} by pressing it against a surface or squeezing it between the gripper jaws. If multiple modalities can be perceived such as when using the BioTac sensors (normal and shear force, vibrations, and temperature) \cite{chu2013using,hoelscher2015evaluation, xu2013tactile} or skin modules (normal force, proximity sensor, acceleration, and temperature) \cite{kaboli2018active}, it becomes possible to discriminate objects based on a combination of these properties. 

The focus of this work is not tactile object recognition as an endeavor exploiting several submodalities of touch (normal and tangential force, vibration, temperature) and involving exploration of the object's surface. Instead, we focus on compressing the object between the fingers of a parallel jaw gripper or a robot hand and using only the mechanical response from squeezing the object---the stress/strain curve---for discrimination. Only one of the devices we used, the Barrett Hand, was equipped with tactile sensors, making it possible to assess their specific contribution.

Stress-strain models were used to gain information about deformable materials without explicitly extracting stiffness/elasticity values in \cite{longhini2023elastic}. Yao et al. \cite{yaoestimating} used the compression action to create three-dimensional stiffness maps of household objects. Chin et al.~\cite{chin2019automated} acquired data during squeezing of the object with a custom-made soft two-finger gripper with pressure and strain sensors, reaching up to 85\% accuracy of discriminating material (paper, metal, or plastic) of 14 ordinary objects.
An underactuated compliant two-finger gripper with 8 tactile sensors on every finger was used in \cite{liarokapis2015unplanned,spiers2016single}. More standard parallel jaw grippers with tactile sensors were used in \cite{guo2021visual,scimeca2019non,wang2022tactual}. LSTM-based methodology for haptic object recognition via tactile and kinesthetic sensing modalities was proposed in~\cite{pastor2021}. They used an underactuated 3-finger gripper to identify 36 objects belonging to either rigid, soft, or in-bag category. None of the above works compared several grippers or action parameters.
Patni et al.~\cite{patni2024online}, using the same 2-finger grippers and object sets as in this work, assessed the ability of the grippers to estimate the elasticity and viscoleasticity parameters of the objects. Furthermore, fitting the Hunt-Crossley model, they showed how grocery items can be online separated based on their material composition.

Multi-finger hands allow for more sophisticated manipulation. Delgado et al.~\cite{delgado2015tactile} used the Shadow hand to grasp deformable objects and computed the deformability ratio to infer the maximum force allowed to be exerted on the object.
Abderrahmane et al. \cite{abderrahmane2018haptic} developed a zero-shot learning algorithm that was capable of recognizing unknown objects as well. 

\subsection{Sensory channels, features, classifiers, actions}

\textbf{Sensory channels.} Devices used for haptic object recognition can be split to those that are capable of multimodal sensing (e.g., \cite{chu2013using, hoelscher2015evaluation, xu2013tactile, kaboli2018active}), tactile arrays which are composed of several sensors perceiving normal force/pressure \cite{bhattacharjee2012haptic,bhattacharjee2018inferring,guo2021visual,liarokapis2015unplanned,scimeca2019non,spiers2016single}, and perceiving the force-displacement (stress-strain) relationship only through the gripper (typically from the motor effort). Tactile sensing from fingertips may be combined with motor positions and currents \cite{kerzel2019neuro}. For parallel jaw grippers, tactile arrays or pressure sensors may be put on gripper jaws \cite{scimeca2019non, guo2021visual,wang2022tactual}. In this work, we compare three devices with only two sensory channels (position and effort) with the Barrett Hand which is heavily sensorized (3 modalities, 107 channels).

\textbf{Sensory features.} Typically, compressing an object  gives rise to a time series of datapoints with a minimum of two channels: a position channel (like distance between gripper jaws) and force channel (resistance perceived by the gripper or pressure sensed by tactile sensors). Other channels might include joint angles or readings from tactile/pressure sensors.

The most common approach in the literature has been to compute features from these time series. There are two distinct events: (i) the moment of contact with the object and (ii) the final position when compression is stopped. In \cite{liarokapis2015unplanned,spiers2016single}, these events were explicitly exploited, using sensor readings only at these instances. Others used features calculated over the whole time series. The features may be physically motivated, such as the change in force/pressure over the compression \cite{xu2013tactile} or max. force, contact area, and contact motion in \cite{bhattacharjee2012haptic}, or more generic (mean, variance, maximum, minimum, range features in \cite{guo2021visual}). For multimodal data like from the BioTac sensors, more complex feature sets can be constructed \cite{chu2013using,hoelscher2015evaluation}. Tatiya et al. \cite{tatiya2020haptic, tatiya2022transferring} converted visual and tactile feedback data from different robots into higher dimensional, shared latent spaces that allowed the transfer of object knowledge across robotic setups.

Dimensionality reduction techniques like Principal Component Analysis (PCA) can be applied on top of the features \cite{bhattacharjee2012haptic, wang2022tactual} or on raw time series data \cite{hoelscher2015evaluation}.  
In this work, we use a set of general features, similar to \cite{Hoffmann_RAS_2014}, as well as raw time series from individual channels on the input.

\textbf{Classifiers.} The most popular classifiers to classify objects based on deformation cues were Support Vector Machines (SVM) \cite{guo2021visual,hoelscher2015evaluation,decherchi2011tactile,liarokapis2015unplanned,wang2022tactual}, $k$-nearest neighbor ($k$-NN) \cite{bhattacharjee2012haptic,bhattacharjee2018inferring,wang2022tactual}, random forests \cite{hoelscher2015evaluation,liarokapis2015unplanned}, na\"ive Bayes \cite{hoelscher2015evaluation,wang2022tactual}, Hidden Markov Models \cite{bhattacharjee2018inferring}, feed-forward neural networks like Multi-Layer Perceptron \cite{kerzel2019neuro,wang2022tactual}, and recurrent neural networks---Jordan type \cite{hosoda2010robust} or, more recently, LSTMs \cite{bednarek2019robotic,bhattacharjee2018inferring,kerzel2019neuro} or transformers (\cite{gao2024transformer} for a review). Recurrent neural networks can be fed directly with time series data, as used in \cite{hosoda2010robust}.
Here we compare performance of $k$-NN over raw data and features, SVM over features, and LSTM over raw data.

\textbf{Exploratory actions.} Leaving sequences of exploratory actions aside (see \cite{kirby2022comparing} for recent work comparing such sequences with a single grasp), the action parameters of a single compression action---its speed and maximum force---are rarely studied. Often, open loop control with a given velocity is applied until a stopping condition (e.g., force threshold or steady state position). The maximum force threshold should be set correctly in order to induce sufficient compression while at the same time not damaging the object (compressing beyond the elastic limit)~\cite{wang2022tactual}. Xu et al.~\cite{xu2013tactile} used a different approach, maintaining a constant force and measuring deformation.
In this work, we study the effects of different compression speeds and finger configurations for multi-finger hands and parallel jaw grippers.

\section{Experimental setup}
\label{sec:setup}

\subsection{Physical object datasets}
\label{sec:methods_objects}
We considered two sets of deformable objects. The ordinary \textit{Objects set} consisted of 9 mostly cuboid objects with a different size and degree of deformability. The polyurethane \textit{Foams set} consisted of 20 polyurethane foam blocks of similar size, along with reference values for elasticity and density provided by the manufacturer. 
The deformation can be regarded as elastic---objects returning to their undeformed shapes once the external force is removed.

\subsubsection{Objects set}
\label{sec:ordinary_objects}
\begin{figure}[htb]
    \centering
    \begin{subfigure}{0.5\textwidth}
        \includegraphics[width=\textwidth]{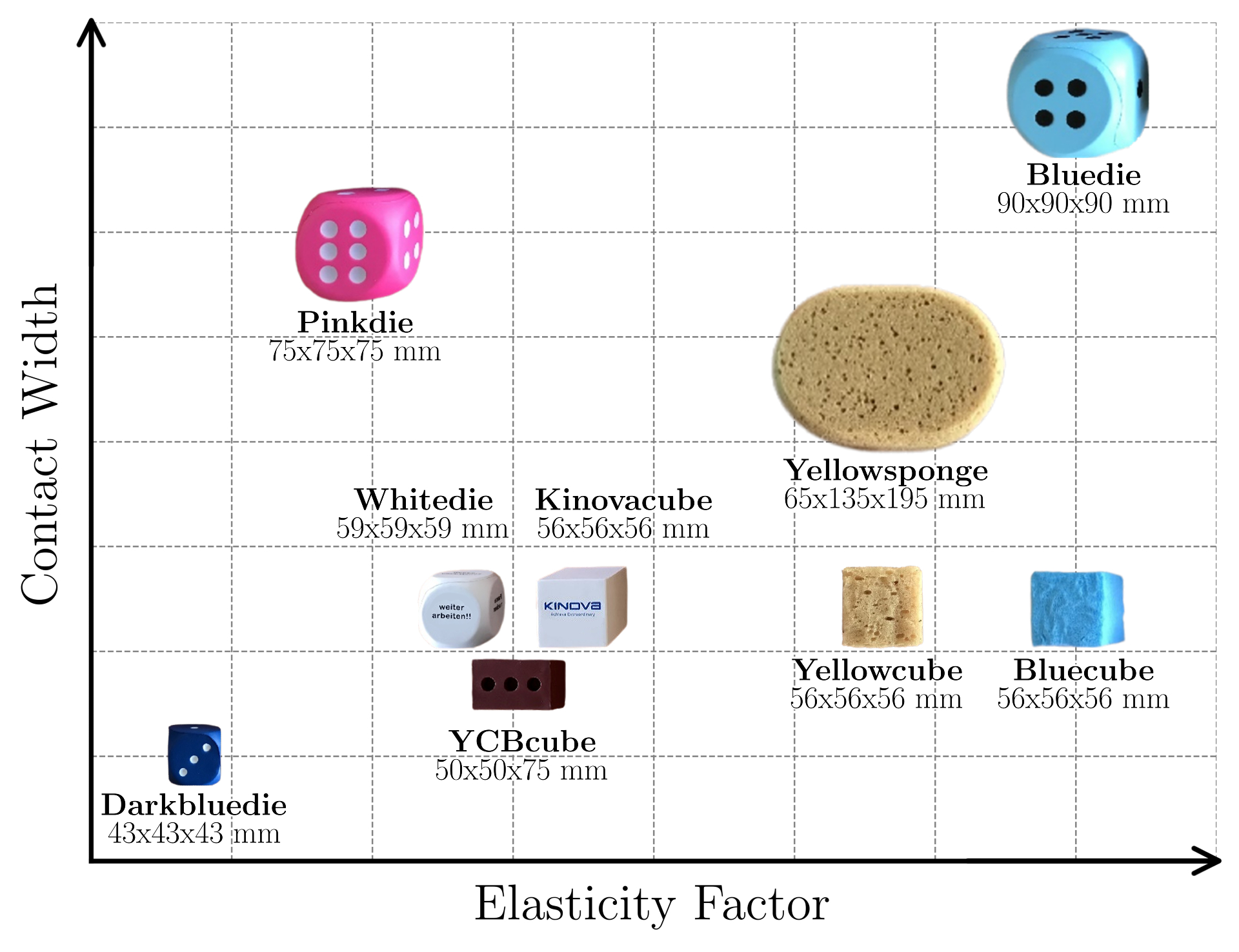}
        \caption{ }
        \label{fig:all_objects}
    \end{subfigure}
    \\
    \begin{subfigure}{0.5\textwidth}
        \centering  % Centers the content within the defined width of the subfigure
        \includegraphics[width=0.7\textwidth]{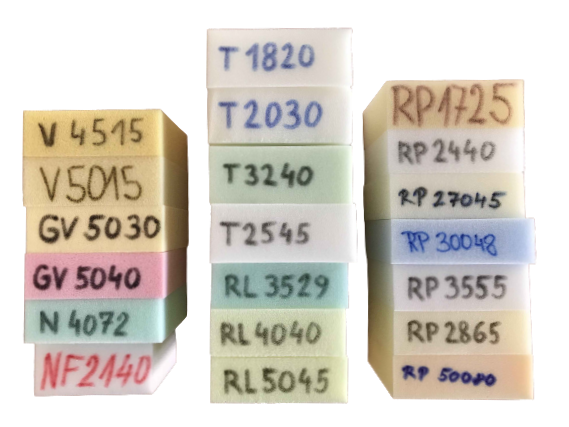}
        \caption{ }
        \label{fig:all_foams}
    \end{subfigure}
    \caption{Physical objects datasets. (a) Ordinary \textit{Objects set} approximately spread out on the elasticity and contact width axes (reference values for this object set are not available). (b)   Polyurethane \textit{Foams set}.}
    \label{fig:objects_and_foams}
\end{figure}

This set of 9 deformable objects is visualized in Fig.~\ref{fig:all_objects}, approximately spread out by the stiffness/elasticity of the objects and the width of the side of the object along which it is pressed by the gripper.
The \emph{yellowcube} is composed of the same material as the \emph{yellowsponge}---it has been cut out from another exemplar of the same sponge, aiming at the dimension of the \emph{kinovacube}. The same is true for \emph{bluedie} and \emph{bluecube}. Conversely, \emph{whitedie}, \emph{kinovacube}, \emph{yellowcube} and \emph{bluecube} have roughly same dimensions but different material composition and hence stiffness. The dataset has been deliberately designed in this way, in order to test which of the object properties are key for model-free haptic object recognition.  

\subsubsection{Polyurethane foams set}
\label{sec:foams}
We used 20 polyurethane foams (Fig.~\ref{fig:all_foams}), samples provided by a manufacturer of mattresses. Their labels encode also reference values for key physical properties: elasticity and density -- see Table~\ref{tab:foams}.

\setlength{\tabcolsep}{4pt}
\begin{table}[htb]
    \centering
    \begin{tabular}{c|c|c|c|c|c}
        Label  & Density & Elasticity & Label  & Density & Elasticity \\
        \hline
        V4515        & 45                                 & 1.5                        &
        V5015        & 50                                 & 1.5                        \\
        GV5030       & 50                                 & 3.0                        &
        GV5040       & 50                                 & 4.0                        \\
        N4072        & 40                                 & 7.2                        &
        NF2140       & 21                                 & 4.0                        \\
        T1820        & 18                                 & 2.0                        &
        T2030        & 20                                 & 3.0                        \\
        T3240        & 32                                 & 4.0                        &
        T2545        & 25                                 & 4.5                        \\
        RL3529       & 35                                 & 2.9                        &
        RL4040       & 40                                 & 4.0                        \\
        RL5045       & 50                                 & 4.5                        &
        RP1725       & 17                                 & 2.5                        \\
        RP2440       & 24                                 & 4.0                        &
        RP27045      & 270                                & 4.5                        \\
        RP30048      & 300                                & 4.8                        &
        RP3555       & 35                                 & 5.5                        \\
        RP2865       & 28                                 & 6.5                        &
        RP50080      & 500                                & 8.0
    \end{tabular}
    \caption{Polyurethane \textit{Foams set}. Label -- code from manufacturer. Density is in ${kg \cdot m^{-3}}$. Elasticity is the $CV_{40}$ in kPa -- ``compression stress value at 40\% strain'' \cite{iso3386}.}
    \label{tab:foams}
\end{table}

\subsection{Robot grippers and hands}
\label{sec:methods_grippers}

\begin{figure}[htb]
    \centering
    \includegraphics[width=0.49\textwidth]{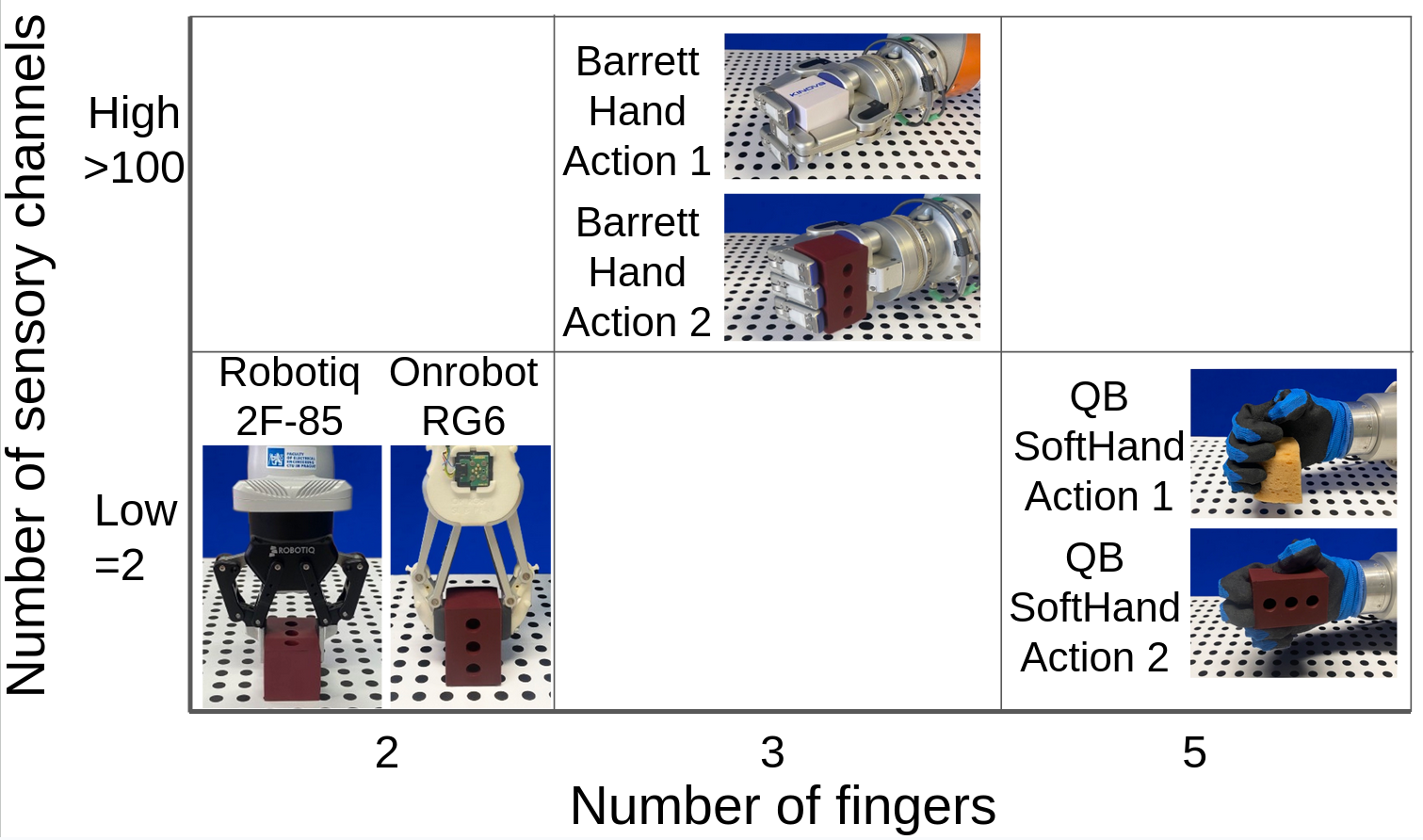}
    \caption{Robot hands and grippers. 
    Two parallel jaw / two-finger grippers were employed: Robotiq 2F-85 and OnRobot RG6, with gripper position and effort (current/force) feedback. The qb SoftHand has five fingers but only one motor and its position and current as feedback. Two action configurations (\emph{a1}: minimizing contact with thumb, \emph{a2}: maximizing contact with thumb) are shown. The Barrett Hand has three fingers that can be rotated around the wrist, 96 tactile sensors, 3 fingertip torque sensors, and 8 joint encoders. We used two finger configurations  (\emph{a1}: opposing fingers, \emph{a2}: lateral fingers). 
    }
    \label{fig:all_hands}
\end{figure}

Four different devices were employed, with different morphology and hence degree of anthropomorphism (industrial type parallel-jaw grippers vs. robot hands with fingers), different options for controlling the squeezing action (compression speed and, if available, gripper configuration), and different type and dimension of feedback---see Fig.~\ref{fig:all_hands} for a schematic overview. 

\textbf{Data collection -- all grippers.} Recording started when the gripper started closing and stopped based on the stopping criterion specified for every gripper separately (detailed below). The time series for subsequent analysis were taken from the moment of contact with the object, which was detected from the effort/current change above a nominal value (5-10\%, depending on the noise level), to the stopping criterion reached. The number of samples collected per object compression was dependent on: (i) the gripper sampling rate, (ii) gripper speed during the experiment, and (iii) the object---for stiffer objects, the stopping criterion was reached earlier, resulting in fewer samples. For the same gripper and speed, the time series were aligned from the end and the shorter ones padded with zeros at the beginning to have the same length. The approximate number of samples collected per gripper and speed is listed below under \textit{Sampling rate and number of samples}.

Below we detail technical parameters of every gripper: the action parameters, the sensory channels, the stopping criterion of the grasping action, and the sampling rate.

\subsubsection{\textbf{qb SoftHand (Research edition)}}
\label{sec:soft_hand}
\begin{figure}[!ht]
\centering
    \begin{subfigure}{0.23\textwidth}
        \includegraphics[width=\textwidth]{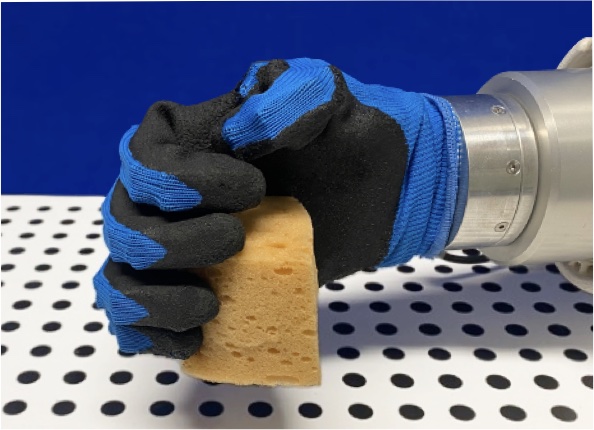}
        \caption{}
        \label{subfig:qb_yellowsponge}
    \end{subfigure}
    \begin{subfigure}{0.23\textwidth}
        \includegraphics[width=\textwidth]{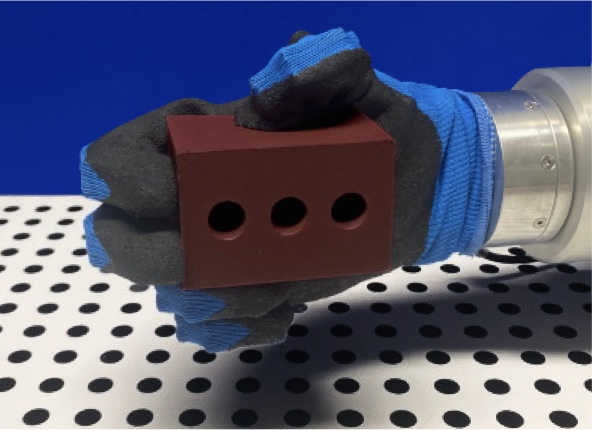}
        \caption{}
        \label{subfig:qb_ycbcube}
    \end{subfigure}
     \begin{subfigure}{0.23\textwidth}
        \includegraphics[width=\textwidth]{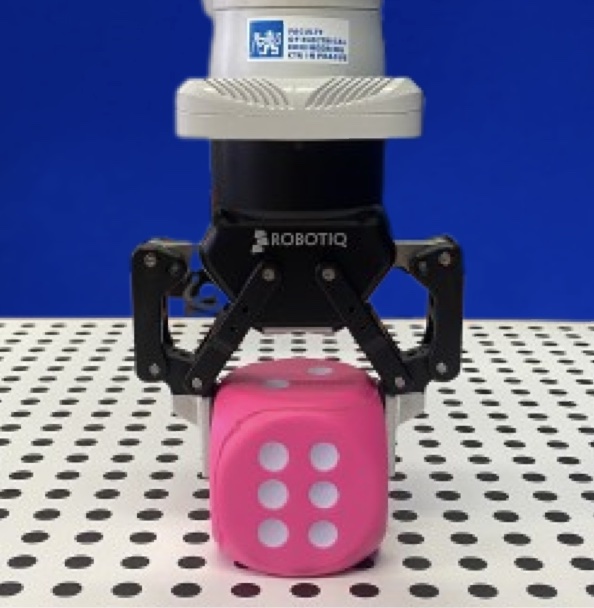}
        \caption{}
        \label{subfig:2f_pinkdie}
    \end{subfigure}
    \begin{subfigure}{0.23\textwidth}
        \includegraphics[width=\textwidth]{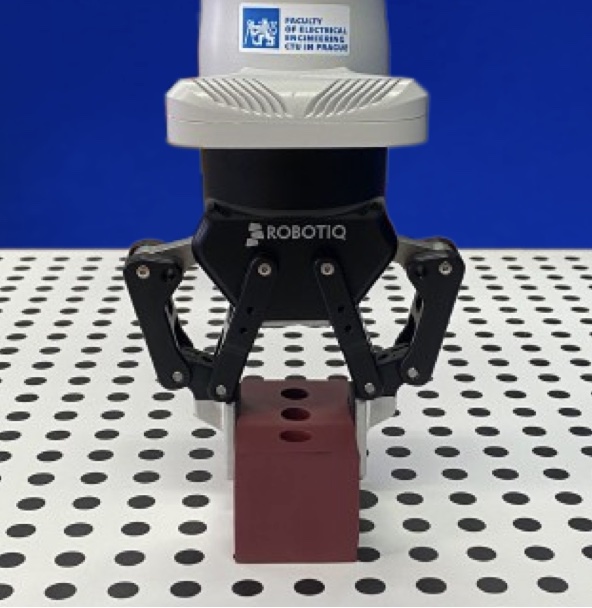}
        \caption{}
        \label{subfig:2f_ycbcube}
    \end{subfigure}
    \begin{subfigure}{0.23\textwidth}
        \includegraphics[width=\textwidth]{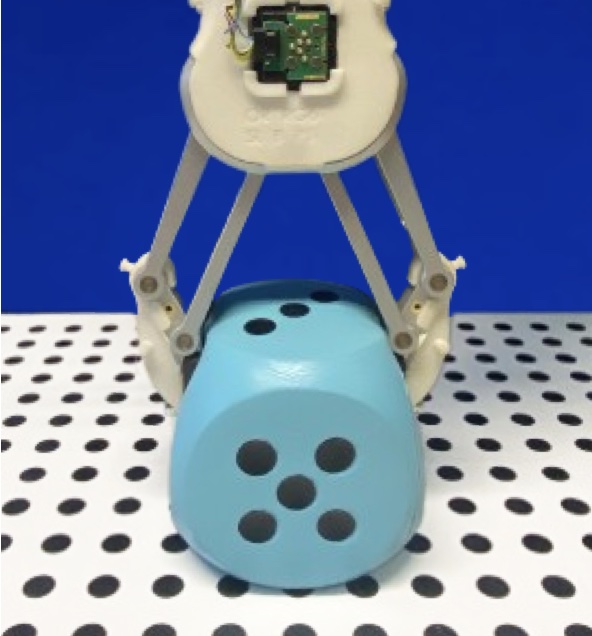}
        \caption{}
        \label{subfig:rg6_bluedie}
    \end{subfigure}
    \begin{subfigure}{0.23\textwidth}
        \includegraphics[width=\textwidth]{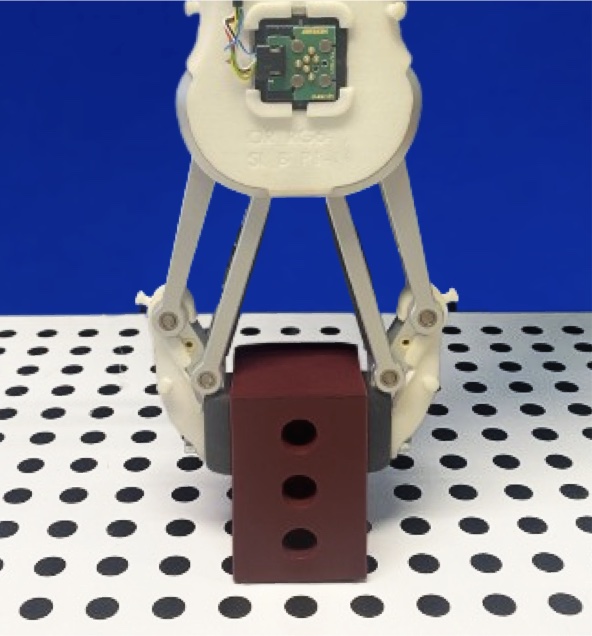}
        \caption{}
        \label{subfig:rg6_ycbcube}
    \end{subfigure}
    \begin{subfigure}{0.23\textwidth}
        \includegraphics[width=\textwidth]{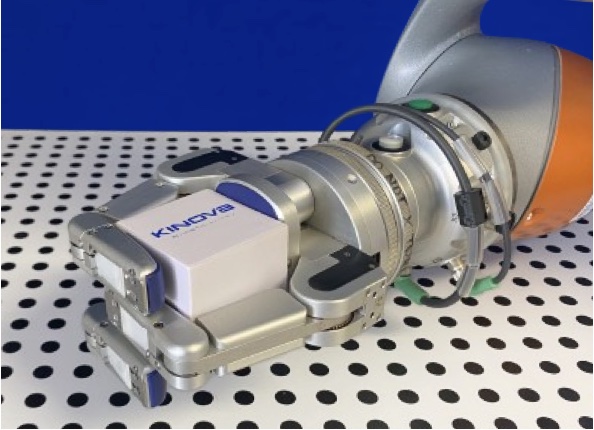}
        \caption{}
        \label{subfig:bh_kinova}
    \end{subfigure}
    \begin{subfigure}{0.23\textwidth}
        \includegraphics[width=\textwidth]{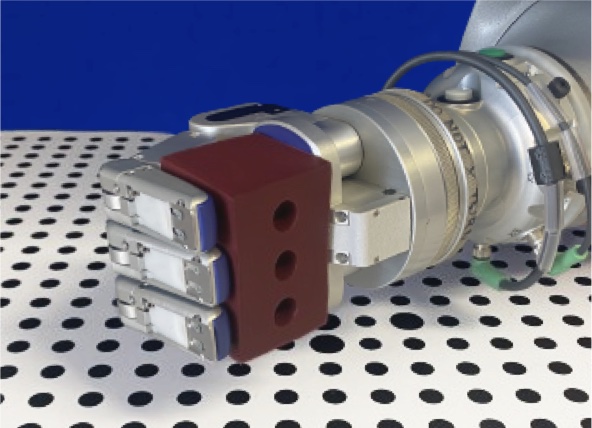}
        \caption{}
        \label{subfig:bh_ycbcube}
    \end{subfigure}
\caption{Experimental setup illustration. (a) qb SoftHand \textit{action1}; (b) qb SoftHand \textit{action2}; (c-d) Robotiq 2F-85; (e-f) OnRobot RG6; (g) Barrett Hand \textit{action1}; (h) Barrett Hand \textit{action2}.}
\label{fig:actions_demo}
\end{figure}

This is an anthropomorphic robotic hand with five fingers controlled via a single electric motor---synchronous opening or closing of all fingers. Due to passive compliance in the joints, it will conform to differently shaped objects. It can reach a grasp force of 62\,N and has a nominal payload of 1.7\,kg. Snapshots from the grasping experiments of two objects with two hand configurations (actions) are shown in Figs.~\ref{subfig:qb_yellowsponge},
\ref{subfig:qb_ycbcube}.

\noindent\textit{Action parameters}: 
We considered two hand configurations or ``actions'': $action1$/\emph{a1} (Fig.~\ref{subfig:qb_yellowsponge}) to minimize contact with the thumb and $action2$/\emph{a2} (Fig.~\ref{subfig:qb_ycbcube}) to maximize it.
The hand was controlled by setting a time when the motor desired position should be reached. We achieved two closing velocities by setting the closing time to 2.5 s ($v1$) and 1.5 s ($v2$).

\noindent\textit{Sensory channels}: 2 channels: motor position and motor current. Since the single motor drives all coupled fingers together, these values indirectly (and ambiguously) code for the position and effort between the fingers and the object.  

\noindent\textit{Stopping criterion}: Using a window of three consecutive samples, the motor positions within this window were compared and if no further movement in the direction of the target motor position was detected, stopping was triggered.

\noindent\textit{Sampling rate and number of samples}: 10 Hz. Each sample contained  28 (17) data points in every channel for $v1$ ($v2$), respectively.

\subsubsection{\textbf{Robotiq 2F-85}}
\label{sec:robotiq_2f85}
The gripper has 2 fingers with 85\,mm stroke with a grip force from 20\,N to 235\,N. The fingertips can be approximated to a rectangle of \begin{math}37.5\times22\,\mathrm{mm}\end{math}. 
Snapshots from the grasping experiments of two objects are shown in Figs.~\ref{subfig:2f_pinkdie},
\ref{subfig:2f_ycbcube}.

\noindent\textit{Action parameters}: Velocity control was used for the closing. This is quoted in \% of the maximum closing speed. Four different nominal closing speeds were executed: 0.68\% ($v0.0068$, approx. 1.6 mm/s), 20.69\% ($v0.21$, 46 mm/s), 63.42\% ($v0.63$, 80 mm/s), and 100\% ($v1.0$, 131.33 mm/s).

\noindent\textit{Sensory channels}:  2 channels: gripper position (gap between jaws 0--85 cm), motor current (A). 

\noindent\textit{Stopping criterion}: The gripper provides feedback of the gap between its jaws as a percentage of the maximum gap. Using a window of eight consecutive samples and a threshold found empirically, the stagnant state, i.e. gap between the jaws remaining constant, was detected and used as a stopping criterion.

\noindent\textit{Sampling rate and number of samples}: 100 Hz. An average sample contained around 760 points ($v0.0068$), 439 ($v0.21$), 356 ($v0.63$), and 319 ($v1.0$) data points in every channel, respectively. 

\subsubsection{\textbf{OnRobot RG6}}
\label{sec:onrobot_rg6}
The gripper has a 160\,mm stroke and adjustable maximum force threshold. The gripper fingertips' surface area is \begin{math} 866\,\mathrm{mm}^{2} \end{math}. Snapshots from the grasping experiments of two objects are shown in Figs.~\ref{subfig:rg6_bluedie},
\ref{subfig:rg6_ycbcube}.

\noindent\textit{Action parameters}:
Force threshold (we used 39N and 109N) and closing velocity. Force thresholds affect the gripper closing speed and were chosen such that the closing speed approximately corresponds to the speeds used on the Barrett Hand (see Section~\ref{sec:velocity_matching} below): 
 $v1=46$ and $v2=93$ mm/s.

\noindent\textit{Sensory channels}: 2 channels: gripper position (gap between jaws), motor current. 

\noindent\textit{Stopping criterion}: Effort (current) threshold reached.

\noindent\textit{Sampling rate and number of samples}: 15 Hz. Each sample contained around 45 (41) points in every channel for $v1$ ($v2$), respectively.

\subsubsection{\textbf{Barrett Hand}}
\label{sec:barrett}
The Barrett Hand (model BH8-282) has three fingers, of which two can rotate around the base.
This design allows numerous configurations of the three fingers, with the only restriction that the spread joints of Finger 1 and 2 are mirroring each other. 

\noindent\textit{Action parameters}: We used two different finger configurations to press the object in the hand: opposing fingers ($action1$ or \emph{a1}; Fig.~\ref{subfig:bh_kinova}) and lateral configuration with all fingers on one side pressing against the palm ($action2$ or \emph{a2}; Fig.~\ref{subfig:bh_ycbcube}). %Three joints, base of every finger,  were controlled during object squeezing. 
Velocity control was used with two different joint velocities: $v1=0.6$, and $v2=1.2$ rad/s. For $action2$ we used forward kinematics to calculate an approximation of the object compression velocity to be 46 and 93 mm/s, respectively.

\noindent\textit{Sensory channels}: 96 pressure-sensitive cells (24 per fingertip and 24 on palm) and 3 fingertip torque sensors. In addition, 8 joint angles from encoders were recorded. 

\noindent\textit{Stopping criterion}: One of three conditions was met: (i) joints not moving anymore, (ii) joint limit reached, (iii) maximum pressure reached.

\noindent\textit{Sampling rate and number of samples}: All data was recorded at 200 Hz, but the effective rate of tactile sensor readings was around 25 Hz. Each sample contained around 260 (220) data points in every channel for $v1$ ($v2$), respectively.

\subsubsection{All grippers -- Compression velocity matching}
\label{sec:velocity_matching}
To the extent that this was possible, the nominal compression velocities (with empty gripper) were chosen such that they were comparable. 
The 39 N threshold on the RG6 results in closing speed of approximately 46 mm/s, which corresponds to 21\% closing speed for the Robotiq 2F-85.

\section{Time series and Classification}
\label{sec:classification_and_unsupervised}

\begin{figure*}[!ht]
\centering
    \begin{subfigure}{0.32\textwidth}
        \includegraphics[width=\textwidth]{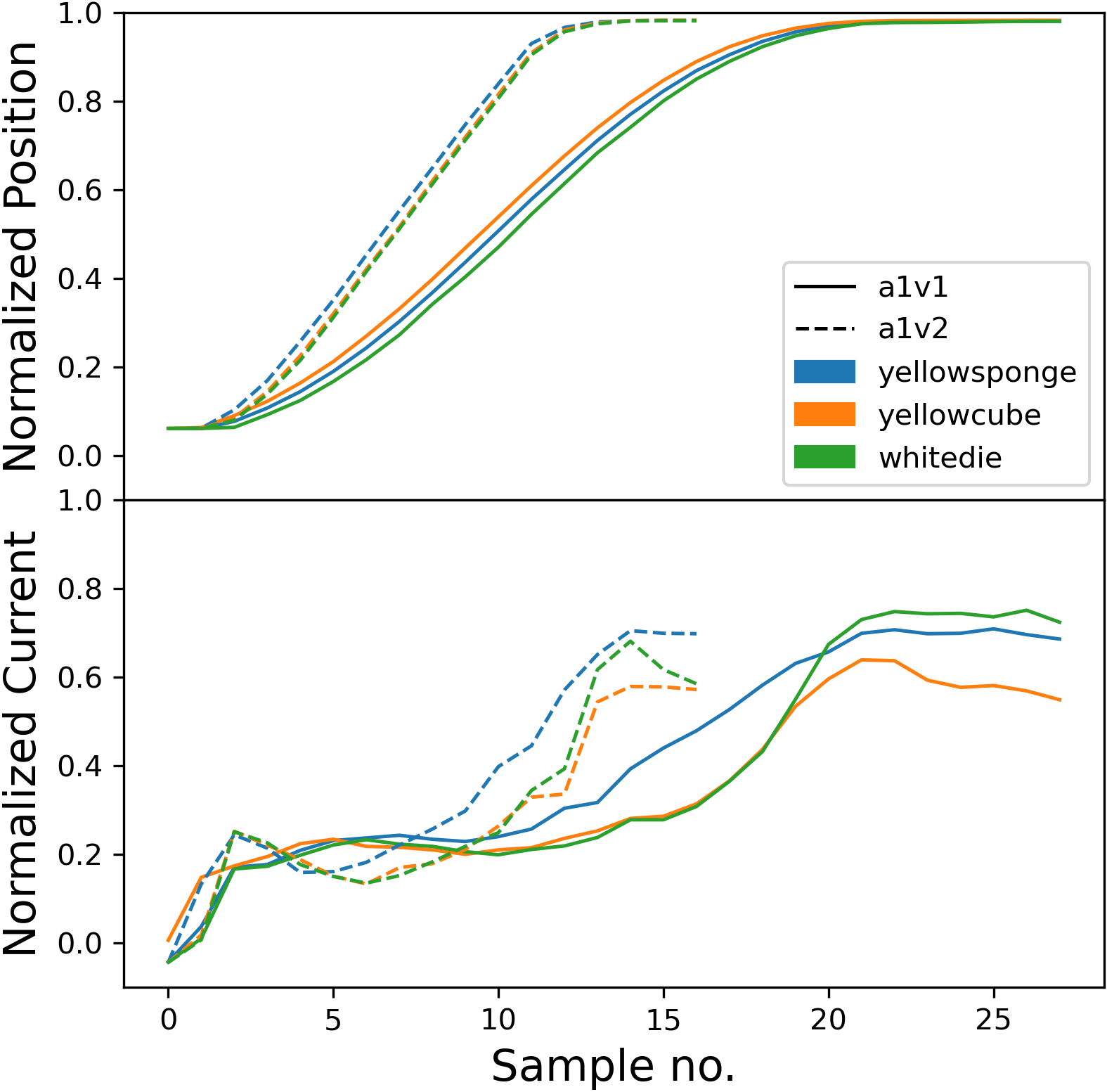}
        \caption{qb SoftHand}
        \label{subfig:raw_qb}
    \end{subfigure}  
    \begin{subfigure}{0.3222\textwidth}
        \includegraphics[width=\textwidth]{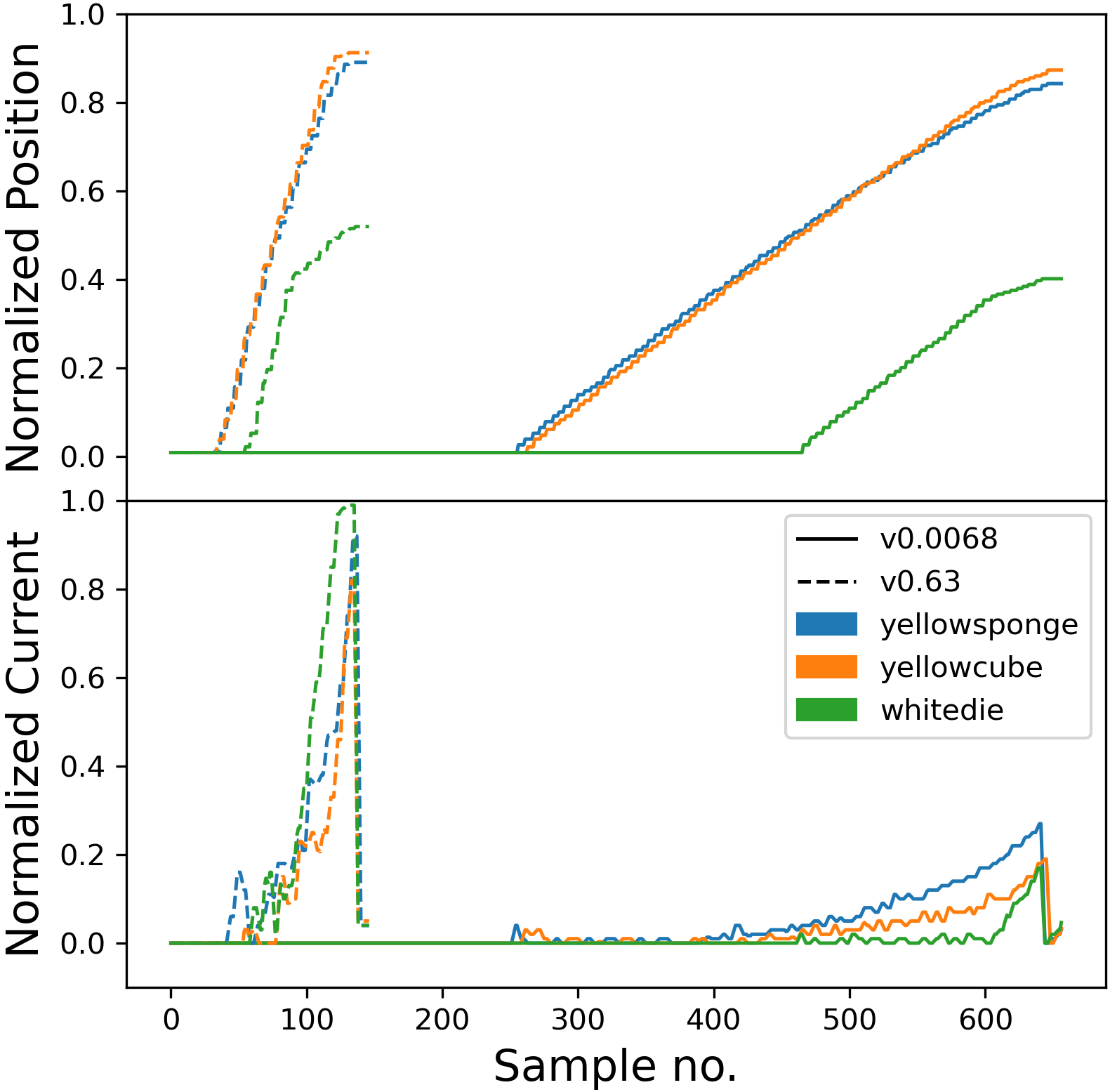}
        \caption{Robotiq 2F-85}
        \label{subfig:raw_robotiq}
    \end{subfigure}
    \begin{subfigure}{0.3222\textwidth}
        \includegraphics[width=\textwidth]{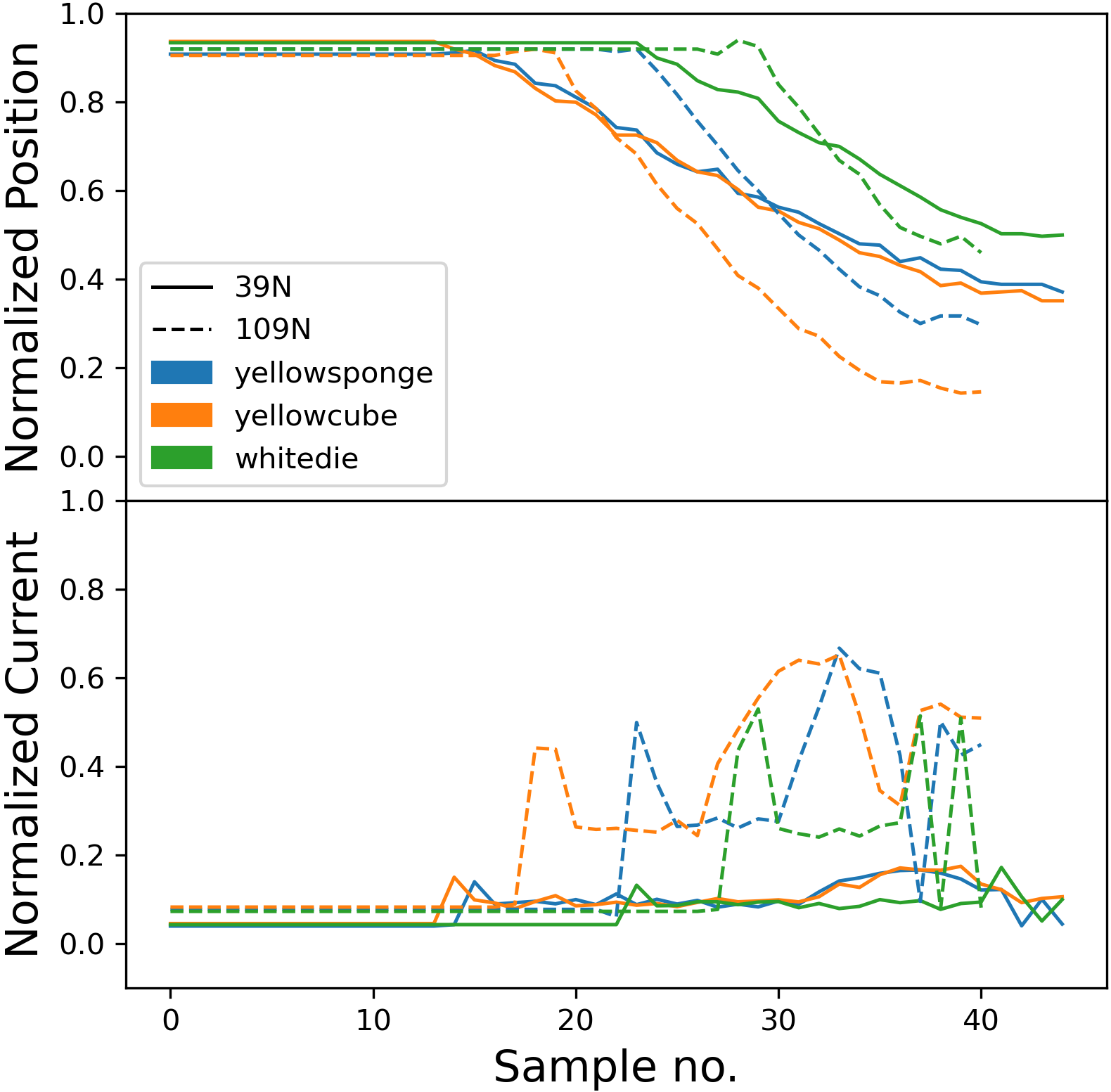}
        \caption{OnRobot RG6}
        \label{subfig:raw_rg6}
    \end{subfigure}  
    \begin{subfigure}{0.32\textwidth}
        \includegraphics[width=\textwidth]{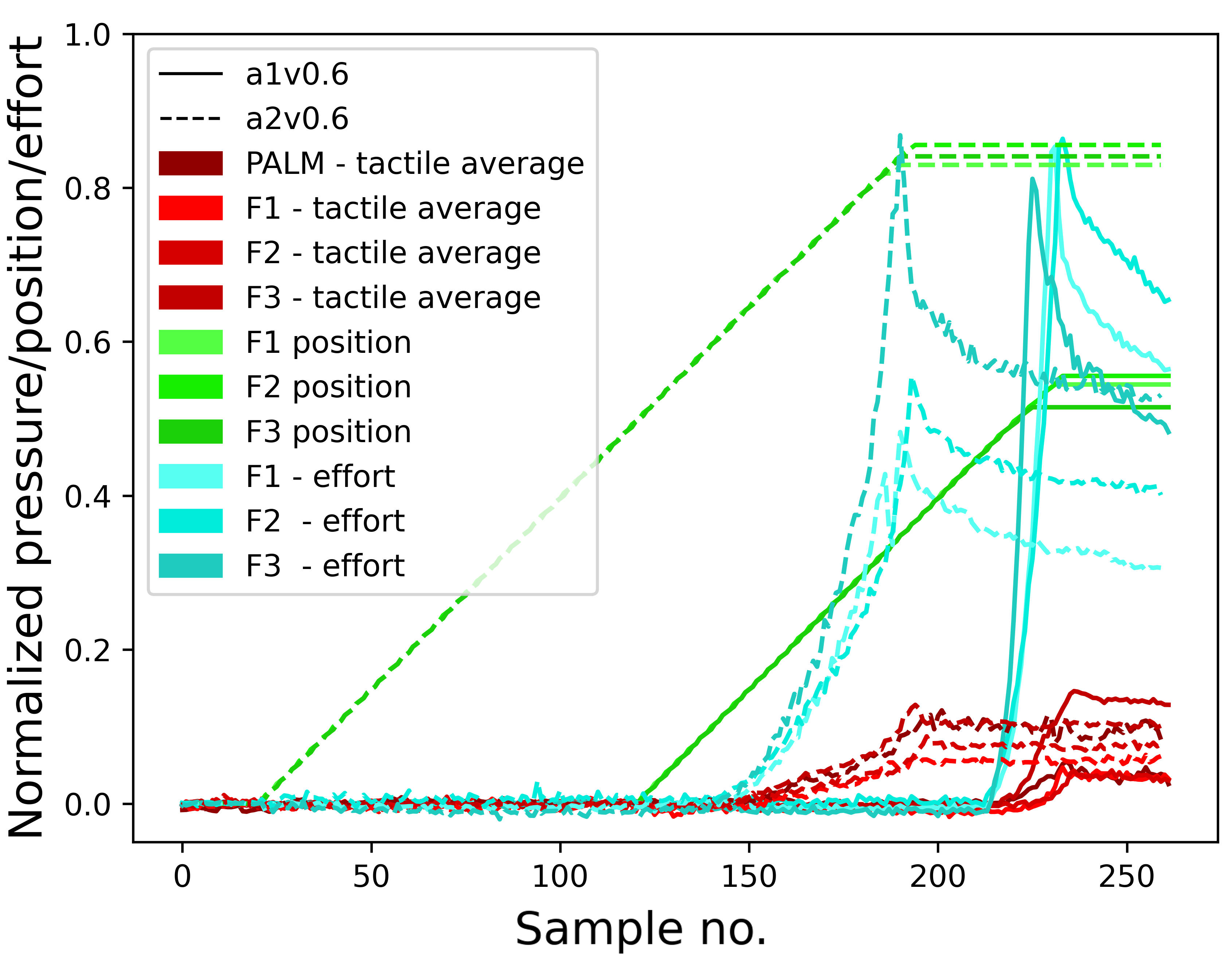}
        \caption{Barrett Hand - yellowsponge}
        \label{subfig:raw_barret-yellowsponge}
    \end{subfigure}
    \begin{subfigure}{0.32\textwidth}
        \includegraphics[width=\textwidth]{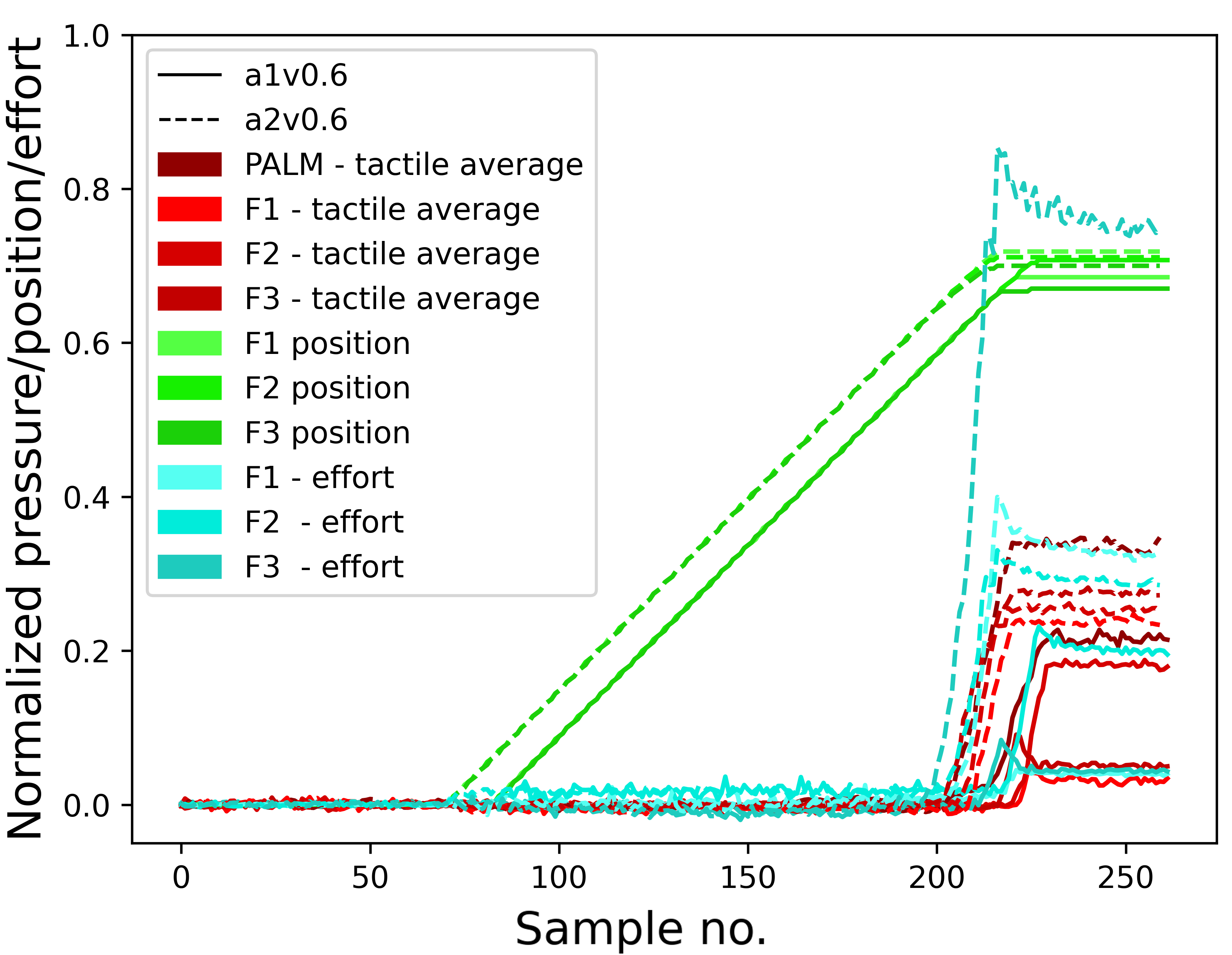}
        \caption{Barrett Hand - whitedie}
        \label{subfig:raw_barret-whitedie}
    \end{subfigure}
    \begin{subfigure}{0.32\textwidth}
        \includegraphics[width=\textwidth]{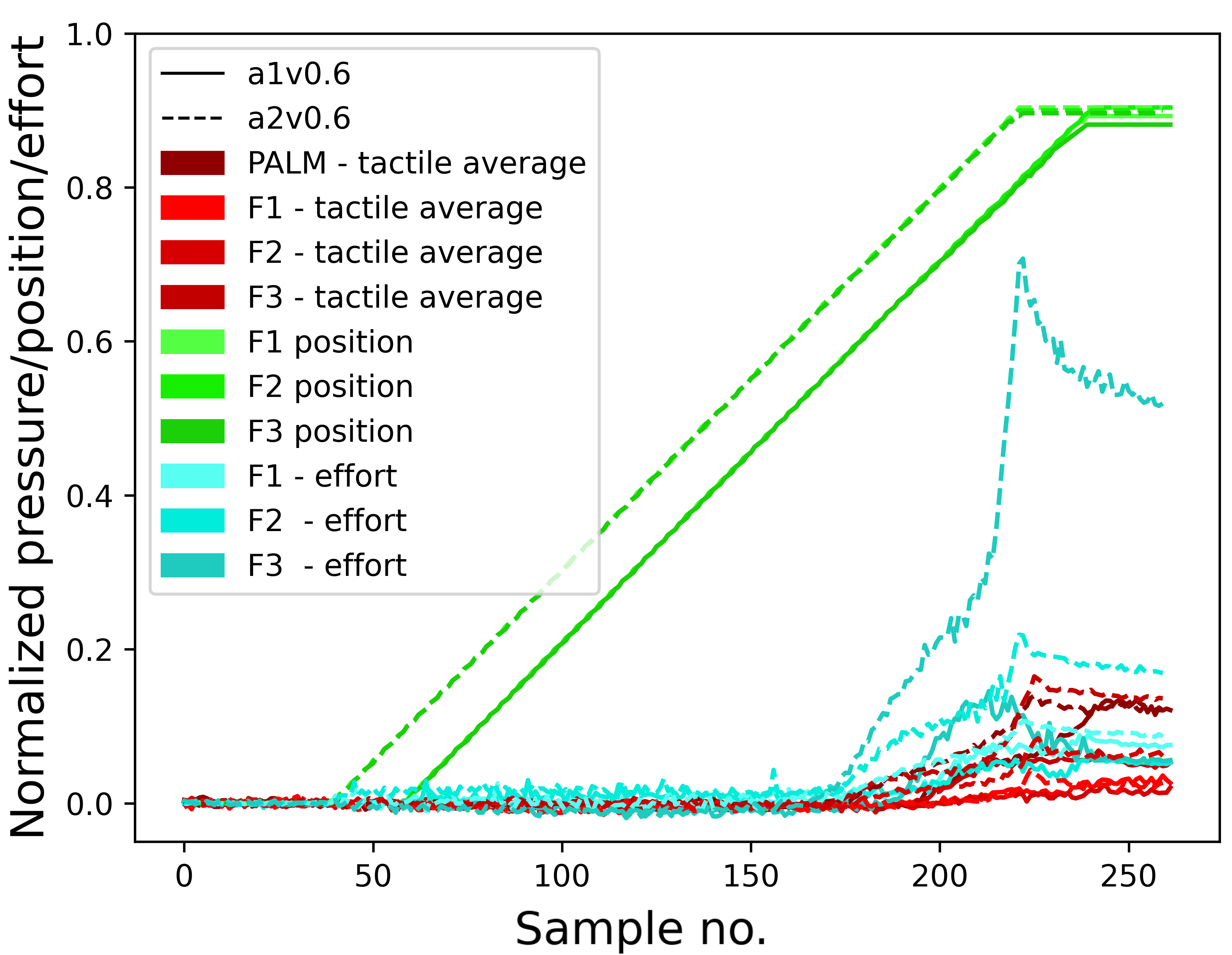}
        \caption{Barrett Hand - yellowcube}
        \label{subfig:raw_barret-yellowcube}
    \end{subfigure}
\caption{Visualization of raw data for individual grippers. 
The effect of action parameters and the grasped objects is illustrated. For the qb SoftHand (a), Robotiq 2F-85 (b), and OnRobot RG6 (c) gripper, we present the normalized current and normalized position during the grasping of three different objects (\emph{yellowsponge}, \emph{yellowcube}, and \emph{whitedie}) using two different velocity settings. For the Barrett Hand (d)-(f), we show the progression of tactile, position, and effort sensors during grasping for two different finger configurations with a joint velocity of $v = 0.6$ rad/s. The \emph{yellowsponge} and \emph{yellowcube} are made of the same material, while the \emph{yellowcube} and \emph{whitedie} have roughly the same size. }
\label{fig:raw_plots}
\end{figure*}

Fig.~\ref{fig:raw_plots} shows examples of raw data from individual grippers for different action parameters and selected objects.

\subsection{Time series data processing and features}
We used (i) raw sensory data signals or (ii) a vector of hand-crafted features as input for classification. Number of data points per time series was the same for all trials (all objects) with a given device and settings (compression speed). If that was not possible---trial was terminated when no further compression was possible (Barrett Hand, Robotiq 2F-85)---the data was padded with zeros from the beginning.

\textbf{Features.} We used features proposed in \cite{Hoffmann_RAS_2014}: minimum, mean, skewness, maximum, kurtosis, median, standard deviation, sum of values, and amplitude of Hilbert transform.
All features were computed for each sensory channel, concatenated in one multidimensional vector, and normalized with the zero mean and unit standard deviation across all dimensions of the feature vector.

\subsection{Classification}
\label{sec:methods_classifiers}

\textit{Problem formulation:}
Consider the space of all possible measurements $X = \mathbb{R}^{s \times n}$, where $s$ represents the number of sensory channels and $n$ the number of measurements (timesteps). Let $Y = \{0,1, \dots, C\}$ $Y = \mathbb{R}$ 
be the space of labels, where $C+1$ is the number of object categories. We are searching for a function $g$, which will map an input $\mathbf{x} \in X$ to a label $y \in Y$: $g: X \to Y$.

We compared the performance of the following classifiers (implementation from Scikit-learn and PyTorch): $k$-NN, SVM, and LSTM.

\subsubsection{$k$-nearest neighbors ($k$-NN)} The $k$-NN classifier was trained either with raw sensory data or hand-crafted features as input. Euclidean metric was used and $k$ selected empirically by evaluating the performance on the validation set. Each gripper had thus its own optimal $k$ on each object set (objects / foams).

\subsubsection{Support Vector Machine (SVM)}
The SVM Classifier was trained with hand-crafted features on the input. We used grid search to find the optimal values for the hyperparameters (kernel, \(C\), and \(\gamma\)) tailored to a specific gripper, action parameters, and object set. For our experiments, we considered three types of kernel: linear, sigmoid, and RBF. The values for \(C\) were logarithmically distributed, ranging from \(10^{-5}\) to \(10^{-3}\), typically exploring 7-10 samples within this range. For the \(\gamma\) values, when applicable, we selected from a logarithmically spaced set between \(10^{-5}\) and \(10^{-3}\), usually considering 7-10 samples. The combination that yielded the best performance for each specific gripper, action parameter, and object set was chosen as the optimal set of hyperparameters.

\begin{table*}
    \centering
    \includegraphics[width=\textwidth]{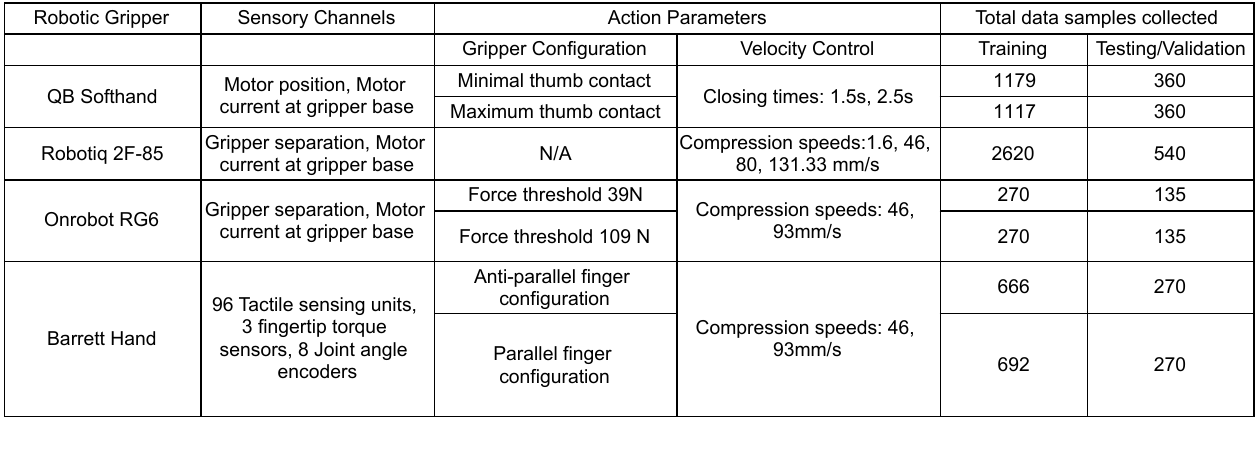}
    \caption{Overview of experiments and settings. One `data sample' is the time series from all sensory channels during a single object compression.}
    \label{tbl:All_experiments}
\end{table*}

\subsubsection{Long-Short Term memory Neural Network (LSTM)}
\label{subsec:methods_lstm}
The LSTM classifier was trained on raw data padded with zeros to the length of the longest measurement, keeping the information about the original length. The output of the last LSTM cell (representing the LSTM-computed features) with nonzero input was selected. The features were then passed into two linear layers to compute the output. Object category was selected using the softmax layer. During training, the network weights were modified. Gradient descent (Adam~\cite{adam2015}) was performed to minimize the cross entropy loss. We used grid search to find the optimal values of the hyperparameters for a given gripper, action parameters, and object set. The search was through the number of LSTM hidden layers (2/4), size of the hidden layer (32/64/128/256), learning rate (log. distribution over [$10^{-5}$, $10^{-3}$]).  
Batch size was determined empirically to comply with the available memory resources.
%The maximum batch size that fitted into memory was used.
This resulted in 336 different LSTM models. The model performing the best on the validation set was selected (typically networks with 4 LSTM layers and 256 neurons in the hidden layer).

\subsubsection{Training, validation, and testing sets}
The datasets used were divided into training, validation and testing subsets. Testing and validation subsets were distinct but had the same size. 
The details of all the experiments done and the dataset compiled are provided in Table~\ref{tbl:All_experiments}.
The `data samples' indicate the number of time series---one time series corresponds to a single object compression. The number of sensory channels depends on the gripper. %\addedk{The column \emph{all} is not always the sum of the individual columns---the number of samples was reduced by using a random selection.}
The data is available at \cite{osf2023single}.

\subsection{Unsupervised analysis of time series data}
\label{subsec:methods_clustering}
In order to understand what the dominant sources of variability in the data are---the different grippers, the objects explored, the way they are manipulated, the sensory channels used---unsupervised learning techniques can be employed. For data in the form of features, standard dimensionality reduction techniques such as Principal Component Analysis (PCA) can be used to visualize the data. More recently, other techniques like Gaussian Process Latent Variable Models \cite{lawrence2006local} or t-SNE \cite{van2008visualizing} were introduced. We experimented with PCA and t-SNE. As the visualizations were qualitatively similar, only results from PCA are shown here. Interested readers can generate their own visualizations using both PCA and t-SNE through the interactive tool we provide at \cite{osf2023single}.

\begin{figure*}[htbp]
\centering
 \begin{tikzpicture}    
 \matrix (fig) [matrix of nodes,
             column sep=-0.3cm, row sep=-0.3cm]{
 \includegraphics[width=0.25\textwidth]{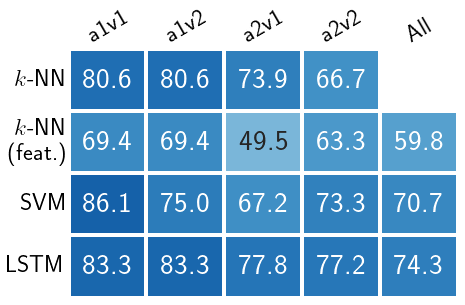}
 &
\includegraphics[width=0.26727273\textwidth]{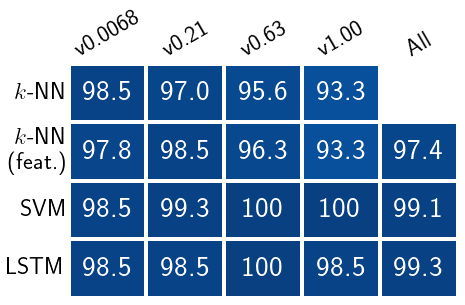}
&
\includegraphics[width=0.173\textwidth]{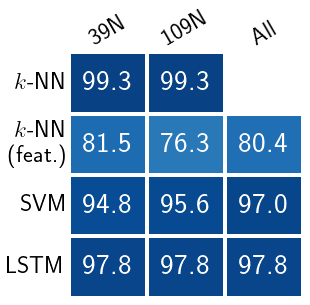}
 &
\includegraphics[width=0.26727273\textwidth]{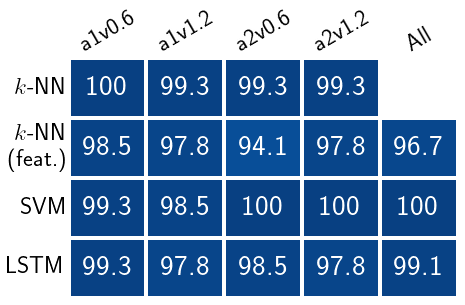}
 \\%[-1cm] %<- adjust this to move the height of the subcaptions up or down
 |[text width=0.25\textwidth]| {\subcaption{qb SoftHand}}
 &
 |[text width=0.28727273\textwidth]| {\subcaption{Robotiq 2F-85}}
&
 |[text width=0.173\textwidth]| {\subcaption{OnRobot RG6}}
 &
 |[text width=0.28727273\textwidth]| {\subcaption{Barrett Hand}}
 \\[0.2cm] %<- adjust this to make the distance to the next row smaller or larger
 &
 \includegraphics[width=.26727273\textwidth]{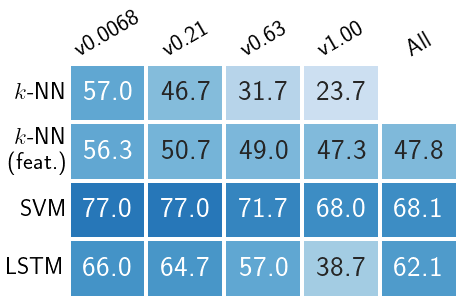}
&
 \includegraphics[width=0.173\textwidth]{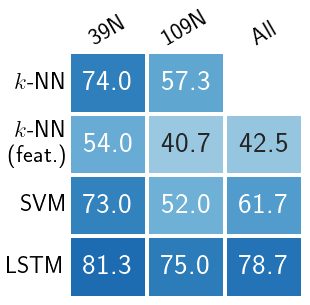}
 &
 \includegraphics[width=0.26727273\textwidth]{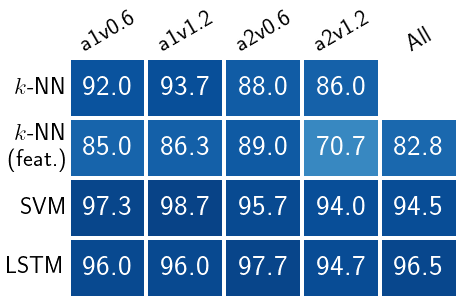}
 \\%[-1cm] %<- adjust this to move the height of the subcaptions up or down
 |[text width=0.25\textwidth]| {}
 &
 |[text width=0.28727273\textwidth]| {\subcaption{Robotiq 2F-85}}
&
 |[text width=0.173\textwidth]| {\subcaption{OnRobot RG6}}
 &
 |[text width=0.28727273\textwidth]| {\subcaption{Barrett Hand}}
 \\
  };
  \path (fig-1-1.south west)  -- (fig-1-1.north west) node[midway,above,sloped]{Objects set};
    \path (fig-4-1.south west)  -- (fig-2-1.north west) node[pos=.35, above,sloped]{Foams set};
 \end{tikzpicture}
    \caption{Classification accuracy -- grippers, classifiers, and action parameters. Subfigures (a)--(d) Objects set; (e)--(g) Polyurethane foams set. Rows represent different classifiers, and columns represent action parameters. Chance level performance: Objects set: 11\%; Foams set: 5\%.} \label{fig:merged_classification}
\end{figure*}

\section{Results -- classification}
\label{sec:results_classification}

\subsection{Ordinary objects set -- classification results}
Time series from the four different hands/grippers compressing the 9 objects were used to train and evaluate the classifiers. An overview of the classification results for the different devices is in Fig.~\ref{fig:merged_classification}, top. The performance of different classifiers is shown in rows in every panel. Overall, the choice of classifier did not play an important role, with $k$-NN on features (\textit{k-NN(feat.)}) performing slightly worse (compared to \cite{wang2022tactual}, where the $k$ -NN outperformed the SVM and LSTM classifiers). Different action parameters---grasp velocity or configuration---are compared in the columns of every panel (see Sec.\ref{sec:action_pars}) for a detailed analysis of the effect of velocity on accuracy). Two of the devices, Robotiq 2F-85 and Barrett Hand, achieved close to perfect performance across action parameters and classifiers. The performance of the OnRobot RG6 gripper is close behind. For selected parameters, Barrett Hand and Robotiq 2F-85 achieved 100\%. 
The qb SoftHand achieved comparatively worse results on the classification tasks---the best performing SVM around 86\% for the slow speed. 

\subsection{Polyurethane foams set -- classification results}
In order to pose a bigger challenge, we tested the Robotiq 2F-85, OnRobot RG6, and the Barrett hand on the set of 20 polyurethane foams (Section~\ref{sec:foams} and Fig.~\ref{fig:all_foams}). Note that this is a highly challenging dataset, impossible for humans to classify correctly. The results are summarized in Fig.~\ref{fig:merged_classification}, bottom. The best results were achieved by the Barrett Hand (between 94\% and 99\% for different actions with the SVM and LSTM classifiers), followed by OnRobot RG6 (over 75\% with LSTM) and Robotiq 2F-85 gripper (between 68\% and 77\% across gripping speeds with SVM classifier). $k$-NN is lagging behind SVM and LSTM.

\begin{figure*}[!htbp]
\centering
 \begin{tikzpicture}    
 \matrix (fig) [matrix of nodes,
             column sep=-0.3cm, row sep=-0.3cm]{
 \includegraphics[width=0.25\textwidth]{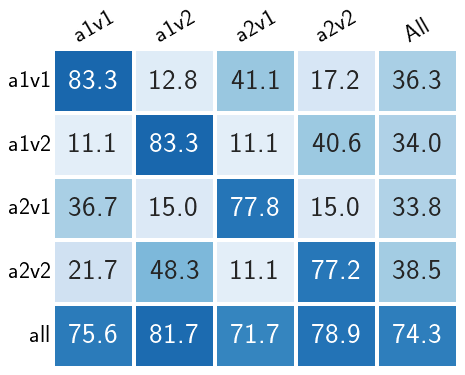}
 &
\includegraphics[width=0.26727273\textwidth]{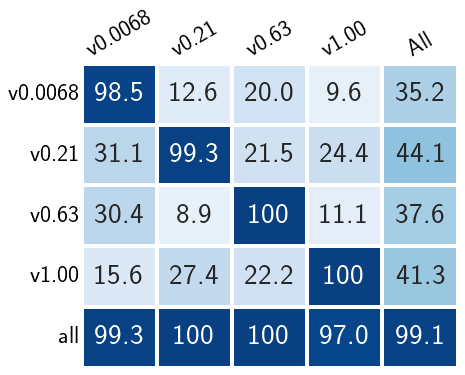}
&
\includegraphics[width=0.173\textwidth]{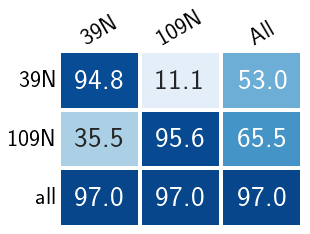}
 &
\includegraphics[width=0.26727273\textwidth]{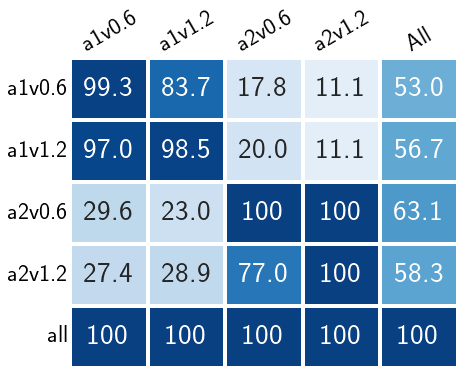}
 \\%[-1cm] %<- adjust this to move the height of the subcaptions up or down
 |[text width=0.25\textwidth]| {\subcaption{qb SoftHand}}
 &
 |[text width=0.28727273\textwidth]| {\subcaption{Robotiq 2F-85}}
&
 |[text width=0.173\textwidth]| {\subcaption{OnRobot RG6}}
 &
 |[text width=0.28727273\textwidth]| {\subcaption{Barrett Hand}}
 \\[0.2cm] %<- adjust this to make the distance to the next row smaller or larger
 &
 \includegraphics[width=.26727273\textwidth]{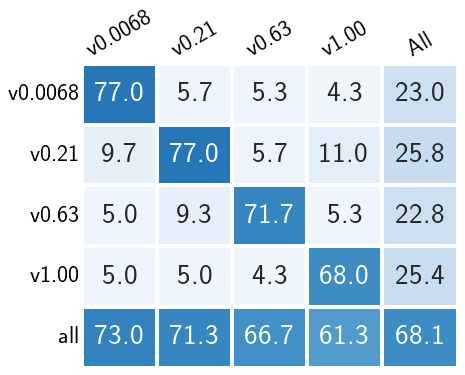}
&
 \includegraphics[width=0.173\textwidth]{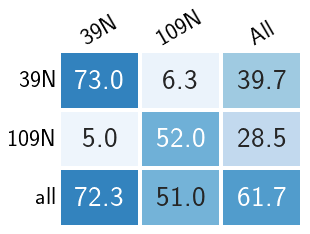}
 &
 \includegraphics[width=0.26727273\textwidth]{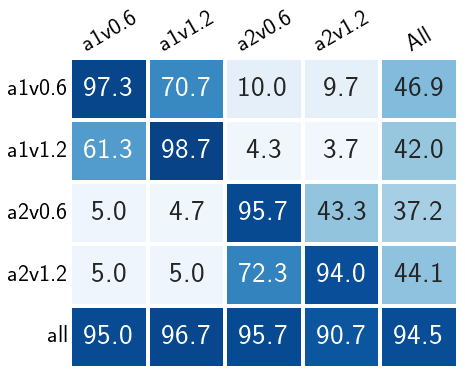}
 \\%[-1cm] %<- adjust this to move the height of the subcaptions up or down
 |[text width=0.25\textwidth]| {}
 &
 |[text width=0.28727273\textwidth]| {\subcaption{Robotiq 2F-85}}
&
 |[text width=0.173\textwidth]| {\subcaption{OnRobot RG6}}
 &
 |[text width=0.28727273\textwidth]| {\subcaption{Barrett Hand}}
 \\
  };
  \path (fig-1-1.south west)  -- (fig-1-1.north west) node[midway,above,sloped]{Objects set};
    \path (fig-4-1.south west)  -- (fig-2-1.north west) node[pos=.35, above,sloped]{Foams set};
 \end{tikzpicture}
    \caption{Generalizing to new action parameters -- classification accuracy for individual grippers when trained on one set of action parameters and tested on another. In individual subfigures, rows represent the training set and columns represent the testing set. Diagonal values indicate results when training and testing were done on the same set of action parameters, corresponding to the results in Fig.~\ref{fig:merged_classification}. SVM classifier used throughout, except for (a) where LSTM was used. Chance level performance: Objects set: 11\%; Foams set: 5\%.}
    \label{fig:class_transfer}
\end{figure*}

\subsection{Effect of action parameters}
\label{sec:action_pars}
First, we studied the effect of speed of object compression. %Robotiq 2F-85, qb SoftHand, and Barrett Hand were used. 
The results are in the columns of Figs.~\ref{fig:merged_classification}(top) and \ref{fig:merged_classification}(bottom). 

We applied up- and downsampling techniques to make the number of data points for different speeds equal.  There is a trend that faster compression speeds degrade classification performance, suggesting that the mechanical response of the objects at slower compression speed facilitates discrimination. 

On the anthropomorphic hands, different configurations of the hand (Barrett Hand) or different placement of the object in the hand (qb SoftHand) were possible---labeled as \emph{a1} or \emph{a2} in Figs.~\ref{fig:merged_classification}(top) and \ref{fig:merged_classification}(bottom). The configuration has an important effect on the qb SoftHand only (\emph{a1} giving better results).

\subsection{Generalization across action parameters}
\label{sec:class_transfer}
Fig.~\ref{fig:class_transfer} provides an overview how classifiers trained on a dataset with certain action parameters---compression speed or finger configuration---generalize to data collected with different parameters. First, if training is performed on all datasets (last row in individual subplots), the performance on individual testing sets is as good as that of classifiers trained on that specific condition (diagonal). Second, generalization across different finger configurations (from \emph{a1} to \emph{a2}) does not work as both the qb SoftHand (panel (a)) and the Barrett Hand (panels d,g) demonstrate. Third,
there is limited generalization to other compression speeds for all the devices used.
On the Barrett Hand (Fig.~\ref{fig:class_transfer}d, ~\ref{fig:class_transfer}g), generalization to other compression speeds is more successful, as the quadrants with the same configuration (\emph{a1}/\emph{a2}) demonstrate.

\subsection{Effect of sensory channels}
\label{sec:class_ablation}
\begin{figure*}[htbp]
\centering
 \begin{tikzpicture}    
 \matrix (fig) [matrix of nodes,
             column sep=-0.3cm, row sep=-0.3cm]{
 \includegraphics[width=0.25\textwidth]{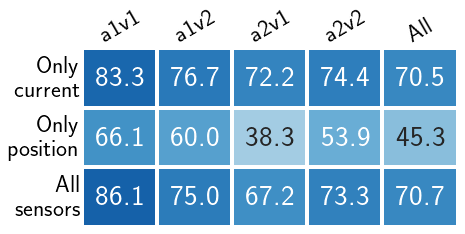}
 &
\includegraphics[width=0.26727273\textwidth]{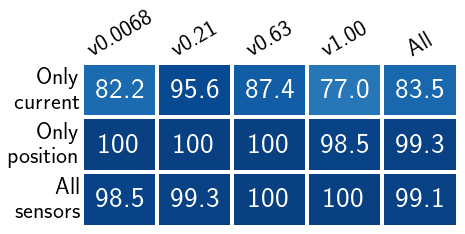}
&
\includegraphics[width=0.173\textwidth]{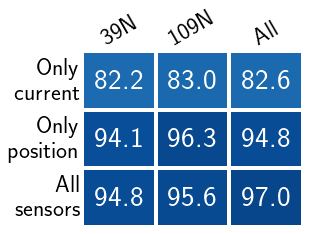}
 &
\includegraphics[width=0.26727273\textwidth]{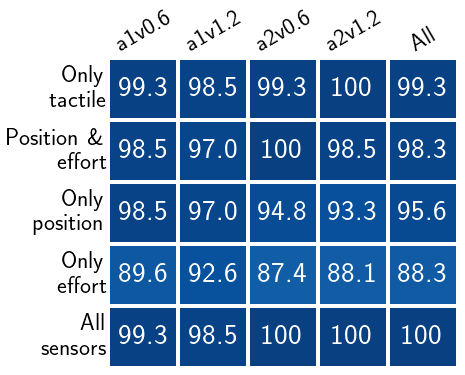}
 \\%[-1cm] %<- adjust this to move the height of the subcaptions up or down
 |[text width=0.25\textwidth]| {\subcaption{qb SoftHand}}
 &
 |[text width=0.28727273\textwidth]| {\subcaption{Robotiq 2F-85}}
&
 |[text width=0.173\textwidth]| {\subcaption{OnRobot RG6}}
 &
 |[text width=0.28727273\textwidth]| {\subcaption{Barrett Hand}}
 \\[0.2cm] %<- adjust this to make the distance to the next row smaller or larger
 &
 \includegraphics[width=.26727273\textwidth]{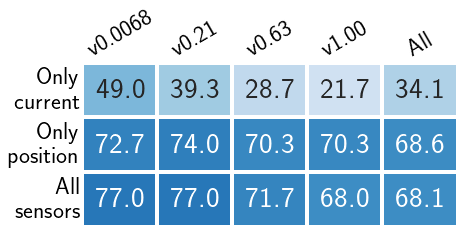}
&
 \includegraphics[width=0.173\textwidth]{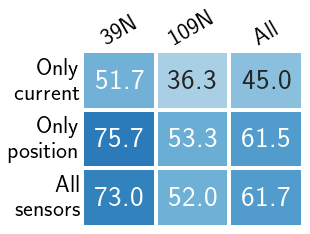}
 &
 \includegraphics[width=0.26727273\textwidth]{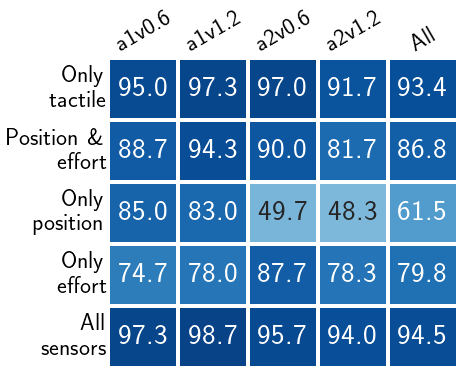}
 \\%[-1cm] %<- adjust this to move the height of the subcaptions up or down
 |[text width=0.25\textwidth]| {}
 &
 |[text width=0.28727273\textwidth]| {\subcaption{Robotiq 2F-85}}
&
 |[text width=0.173\textwidth]| {\subcaption{OnRobot RG6}}
 &
 |[text width=0.28727273\textwidth]| {\subcaption{Barrett Hand}}
 \\
  };
  \path (fig-1-1.south west)  -- (fig-1-1.north west) node[midway,above,sloped]{Objects set};
    \path (fig-4-1.south west)  -- (fig-2-1.north west) node[pos=.35, above,sloped]{Foams set};
 \end{tikzpicture}
    \caption{Effect of sensory channels -- classification accuracy for individual grippers with respect to individual sensors (rows) and employed actions (columns). SVM classifier used in all cases. The chance level performance for the objects set is 11\%; and for the foams set, it is 5\%.}
    \label{fig:class_ablation}
\end{figure*}

We studied which sensory channels are responsible for the classification performance, as shown in Fig.~\ref{fig:class_ablation}.
The performance on the ordinary \textit{objects set} is shown for the qb SoftHand in Fig.~\ref{fig:class_ablation}a, the Robotiq 2F-85 in Fig.~\ref{fig:class_ablation}b, the OnRobot RG6 gripper in Fig.~\ref{fig:class_ablation}c, and the Barrett Hand in Fig.~\ref{fig:class_ablation}d. For the parallel jaw grippers  (Fig.~\ref{fig:class_ablation}b, \ref{fig:class_ablation}c), the position channel alone gives very good performance, similar to position and current (effort) together. 
The current sensor alone performed worse for both grippers. For the qb SoftHand on the \textit{objects set}, Fig.~\ref{fig:class_ablation}a, the pattern is different than for the parallel jaw grippers---the current sensor is performing better than the position sensor. On the Barrett Hand, Fig.~\ref{fig:class_ablation}d, position and effort sensors together or tactile sensors alone provided excellent performance. The fingertip torque sensors alone (effort) had worse performance. 

The performance on the \textit{foams set} is shown in the bottom panels of Fig.~\ref{fig:class_ablation}. The pattern for the 2-finger grippers (Robotiq 2F-85 in (e); OnRobot RG6 in (f)) is similar. Interestingly, position alone has similar performance to position and current together. On the Barrett Hand, (g), the tactile sensors stand out as they are the only modality that alone approaches the performance of all the sensors together.

\section{Results -- Unsupervised analysis}
\label{sec:results_unsup}
In order to understand what the dominant sources of variability in the data are---the different objects explored, the way they are manipulated, the sensory channels used---we employed Principal Component Analysis (PCA). Due to space limitations, only selected six plots are shown and interpreted below. Additional visualizations are available at \cite{osf2023single} (Visualizations using PCA). An interactive tool featuring both PCA and t-SNE is available as well. 

\subsection{Variability across grippers}
\begin{figure*}[h!]
    \centering
    \begin{subfigure}{0.41\textwidth}
        \includegraphics[width=\textwidth]{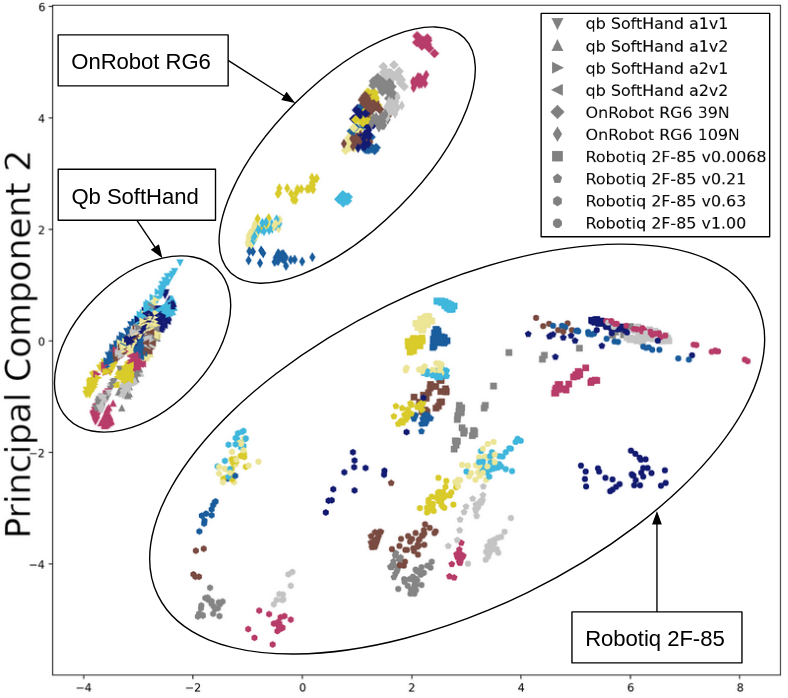}
        \caption{qb SoftHand, Robotiq 2F-85, OnRobot RG6 -- Objects set}
        \label{subfig:PCA_objects_all}
    \end{subfigure}
    \begin{subfigure}{0.4\textwidth}
        \includegraphics[width=\textwidth]{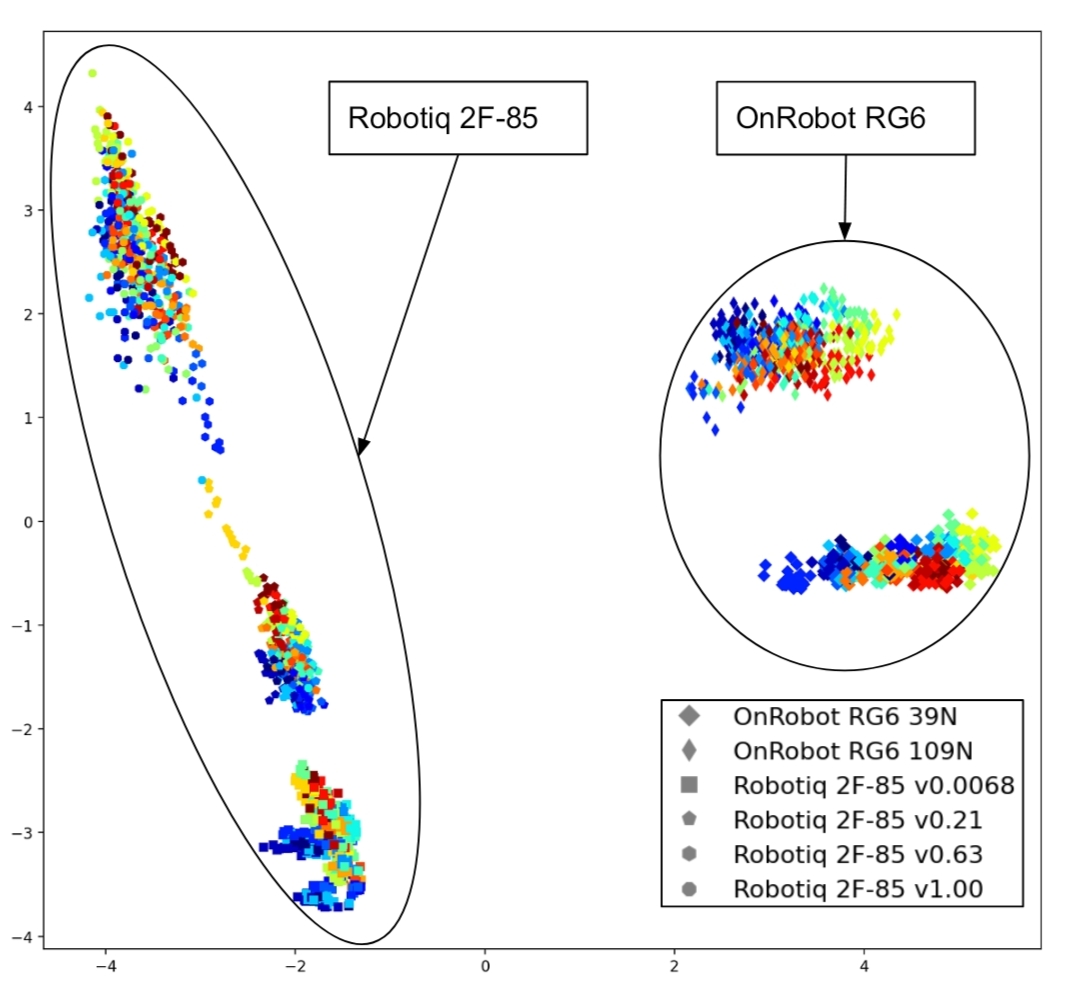}
        \caption{Robotiq 2F-85, OnRobot RG6 -- Polyurethane foams set}
        \label{subfig:PCA_foams_all}
    \end{subfigure}
    \begin{subfigure}{0.17\textwidth}
        \includegraphics[width=\textwidth]{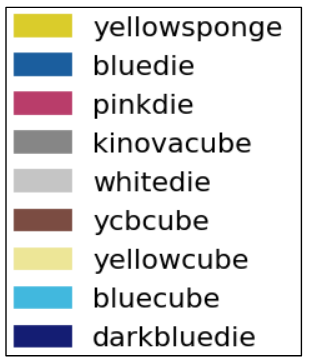}
        \vspace*{.3in}~\\
        \caption*{Legend for the objects set.}
        \label{subfig:objects_legend}
    \end{subfigure}
    \begin{subfigure}{0.41\textwidth}
        \includegraphics[width=\textwidth]{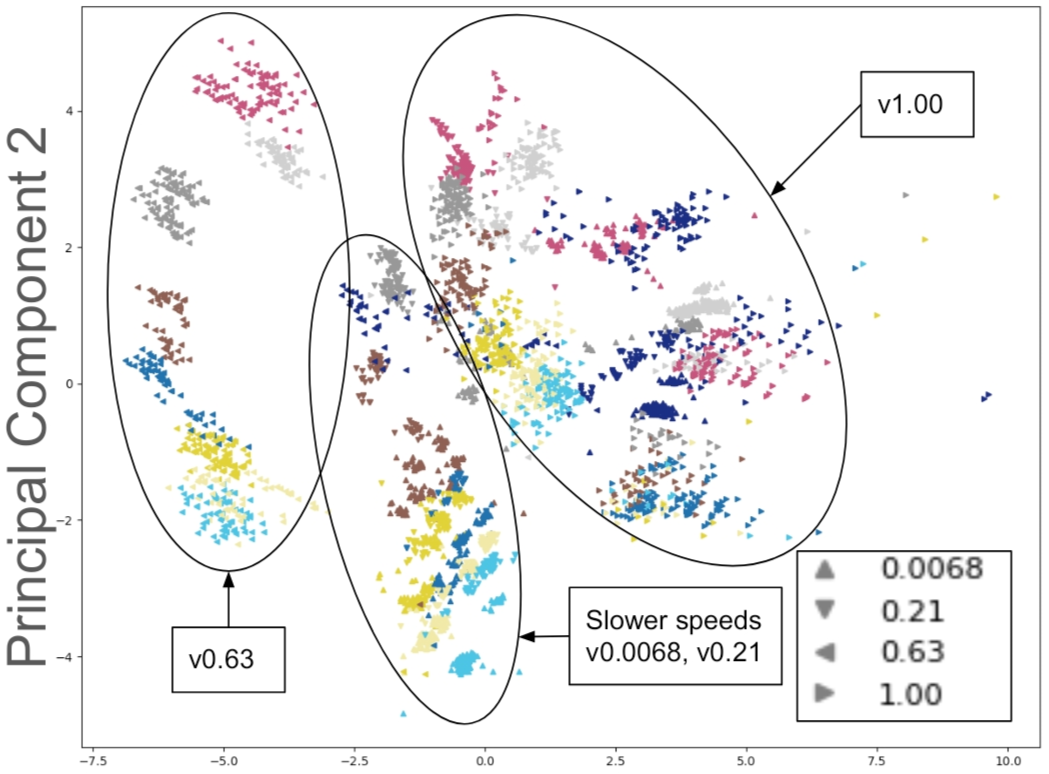}
        \caption{Robotiq 2F-85 -- Objects set}
        \label{subfig:PCA_objects_2F85}
    \end{subfigure}
    \begin{subfigure}{0.4\textwidth}
        \includegraphics[width=\textwidth]{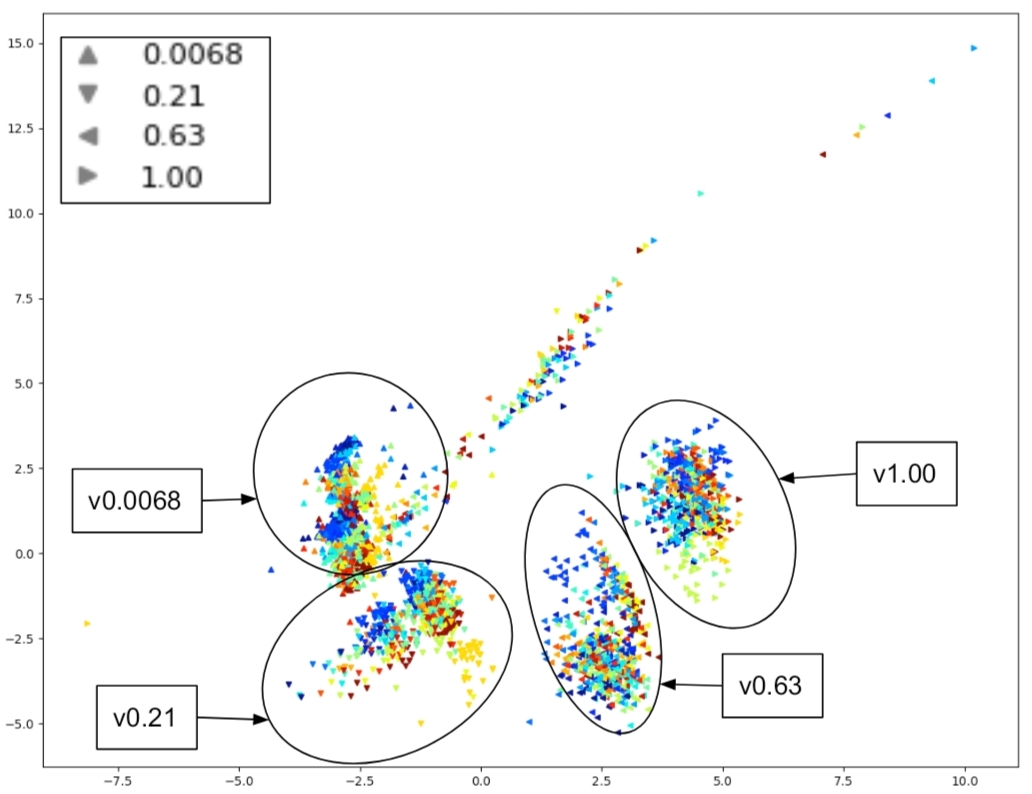}
        \caption{Robotiq 2F85 -- Polyurethane foams set}
        \label{subfig:PCA_foams_2F85}
    \end{subfigure}
    \begin{subfigure}{0.16\textwidth}
        \includegraphics[width=\textwidth]{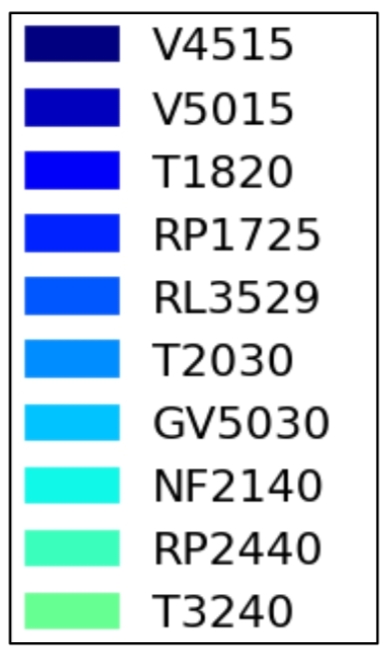}
        \caption*{Legend for the foams set.}
        \label{subfig:foams_legend_1}
    \end{subfigure}
    \begin{subfigure}{0.41\textwidth} %Revert to 0.48
        \includegraphics[width=\textwidth]{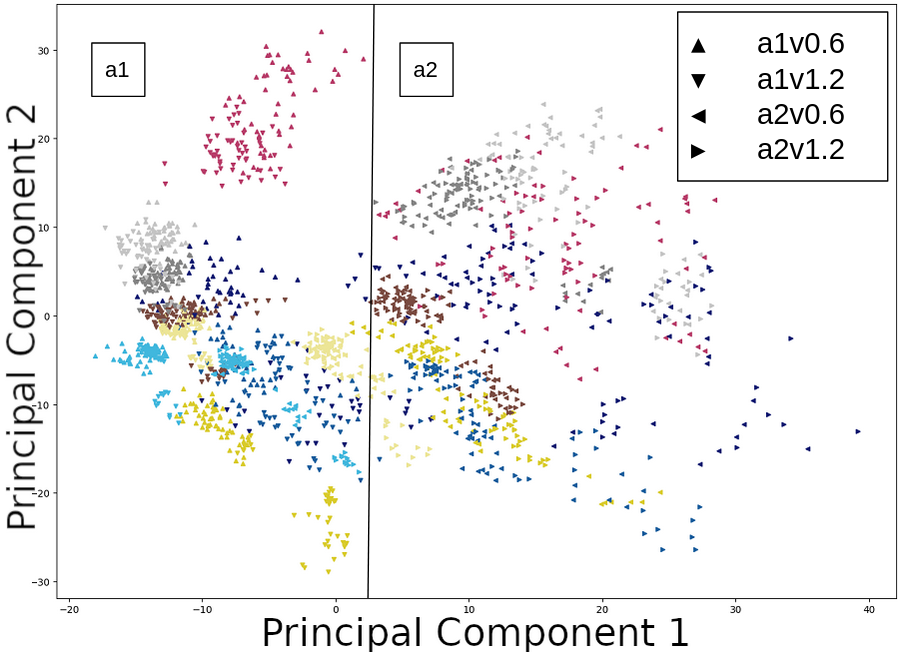}
        \caption{Barrett Hand -- Objects set}
        \label{subfig:PCA_objects_Barrett}
    \end{subfigure}
    \begin{subfigure}{0.4\textwidth} %Revert to 0.47
        \includegraphics[width=\textwidth]{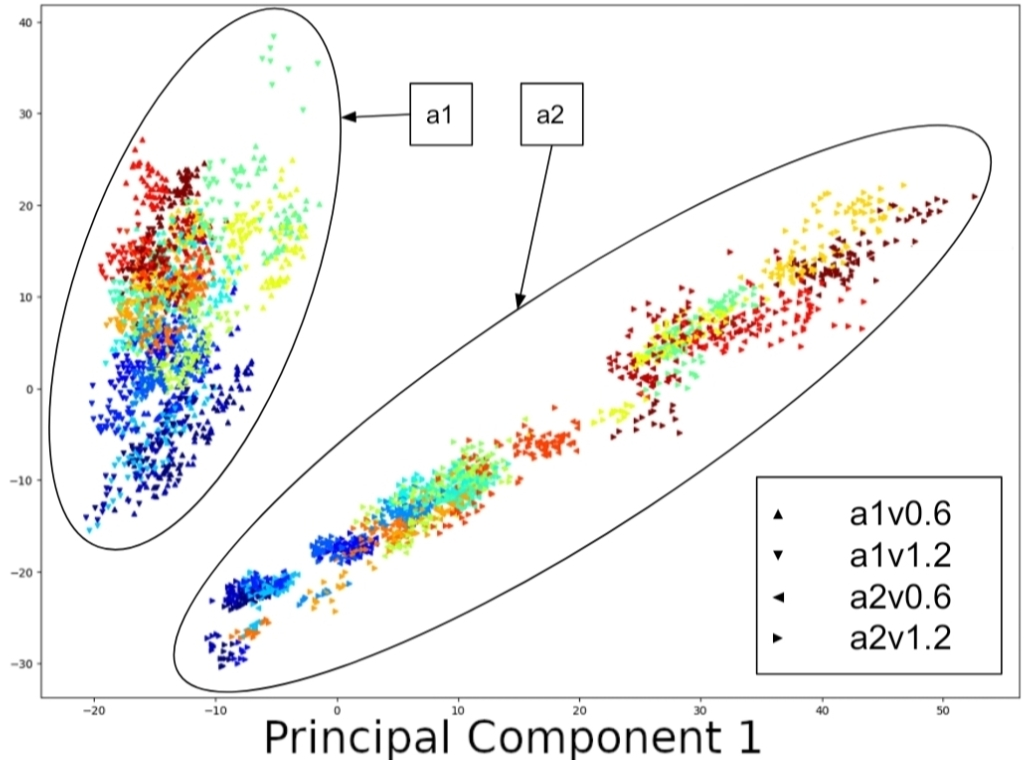}
        \caption{Barrett Hand -- Polyurethane foams set}
        \label{subfig:PCA_foams_Barrett}
    \end{subfigure}
    \begin{subfigure}{0.16\textwidth}
        \includegraphics[width=\textwidth]{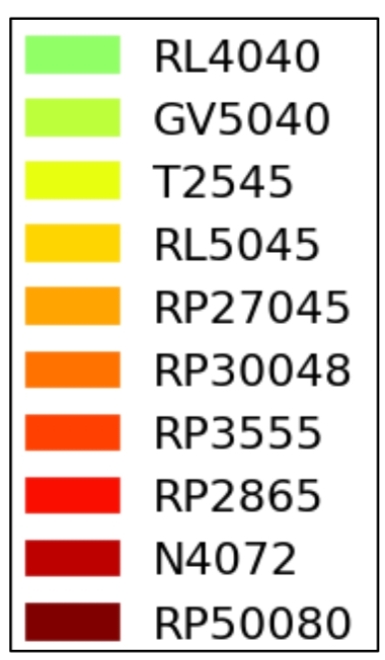}
        \caption*{Legend for the foams set.}
        \label{subfig:foams_legend_2}
    \end{subfigure}
    \caption{Comparing sources of variability using PCA -- different grippers and/or action parameters (markers) and objects grasped (colors). Subfigures a) and b) illustrate whether we can distinguish the data obtained from different grippers for the object sets and foam sets, respectively. Subfigures d) and e) show whether data obtained at different closing speeds can be distinguished when using the Robotiq gripper, for the objects and foam sets, respectively. Finally, subfigures g) and h) display the separation of individual action parameters when using the Barrett hand for the objects and foam sets, respectively.} 
    \label{fig:unsup}
\end{figure*}

First, we assessed the effect of gripper morphology on the variability in the sensory data. Only data (sensory features) from the devices with two sensory channels---OnRobot RG6, Robotiq 2F-85, and qb SoftHand---could be analyzed together. Fig.~\ref{subfig:PCA_objects_all} shows the first two principal components (PC) on the \textit{objects set}. As expected, the different devices shown by different markers form the principal clusters. The qb SoftHand, across configurations and speeds, forms a compact cluster on the left. The OnRobot RG6 gripper occupies the upper part (mainly PC2), with subclusters according to the action parameters. The Robotiq 2F-85 data occupies the bottom central part, with subclusters corresponding to the compression velocities. Interestingly, the different objects (colors) are best separable for the Robotiq 2F-85 gripper, which matches with the classification accuracy.

A similar pattern can be seen on the \textit{foams set} in Fig.~\ref{subfig:PCA_foams_all}, with only the two parallel jaw grippers. The first PC clearly separates the two devices. Interestingly, the second component separates the closing speeds, preserving their order (higher speeds on the top). 

\subsection{Robotiq 2F-85 gripper} 
On the \textit{objects set} (Fig.~\ref{subfig:PCA_objects_2F85}), the first PC is dominated by the compression velocity, with the two slow speeds grouped together. The second PC is then taken up by the object characteristics, dominated by their stiffness. On the \textit{foams set}, Fig.~\ref{subfig:PCA_foams_2F85}, the velocity dominates forming the principal clusters. Note also that the slower speeds better separate the foams. 

\subsection{Barrett Hand} 
On the \textit{objects set}, the first principal component is taken mainly by the grasping configuration with the opposing fingers (\emph{a1}) on the left side of Fig.~\ref{subfig:PCA_objects_Barrett}. Material elasticity seems to dominate PC2 with the softer objects at the bottom. The compression velocity appears to be the smallest source of variability. Unlike on the \textit{objects set}, on the \textit{foams set} visualized in Fig.~\ref{subfig:PCA_foams_Barrett}, two distinct principal clusters are visible corresponding to the grasping finger configuration.  

\subsection{Unsupervised analysis -- summary} For every gripper individually, the first principal component was dominated by the action parameters---the compression speed for OnRobot RG6 gripper, qb SoftHand (not shown here but available at \cite{osf2023single}), Robotiq 2F85, and by grasping configuration for Barrett Hand (for Barrett Hand the compression velocity seems to be the smallest source of variability). The second component is connected to object characteristics---stiffness in the case of OnRobot RG6 gripper, Robotiq 2F85, and Barrett hand, and volume of the objects for qb SoftHand. Similar conclusions can be drawn for the \textit{foams set}. Note also that for the Robotiq 2F-85 gripper, the slower speeds better separate the foams. 
Finally, when sensory data from multiple grippers are combined together (Figs.~\ref{subfig:PCA_objects_all}, \ref{subfig:PCA_foams_all}), it is clearly the morphology or embodiment of the devices that is the main source of variability.

\section{Conclusion and Discussion}
\label{sec:conclusion}

We studied the discrimination of deformable objects by grasping them using 4 different robot hands / grippers: qb SoftHand (5 fingers, 1 motor, position and current feedback), two industrial-type parallel jaw grippers with position and effort feedback (Robotiq 2F-85 and OnRobot RG6), and the Barrett hand (3 fingers with adjustable configuration, 96 tactile, 8 position, 3 torque sensors). A set of 9 ordinary objects differing in size and stiffness and another highly challenging set of 20 polyurethane foams differing in material properties only were used. We systematically compared the grippers' performance, together with the effects of: (1) type of classifier ($k$-NN, SVM, LSTM) operating on raw time series or on features, (2) action parameters (grasping configuration and speed of squeezing), and (3) contribution of sensory modalities. We found: (i) all the grippers but the qb SoftHand could reliably distinguish the ordinary \textit{objects set}; (ii) Barrett Hand reached around 95\% accuracy on the foams; Robotiq 2F-85 around 70\%; (iii) across all grippers, SVM over features and LSTM on raw time series performed best; (iv) faster compression speeds degrade classification performance; (v) transfer learning between compression speeds worked well for the Barrett Hand only; transfer between grasping configurations is limited; (vi) ablation experiments provided intriguing insights---sometimes a single sensory channel suffices for discrimination. Overall, the Barrett Hand as the most complex device with rich sensory feedback provided the best results. The 96 tactile sensors were found to be the most informative sensory modality, compared to the 3 fingertip torque sensors and 8 joint encoders. However, depending on the problem difficulty (\textit{objects set} vs. \textit{foams set}), uncalibrated parallel jaw grippers without tactile sensors can have sufficient performance for single-grasp object discrimination based only on position and effort data only. 

Classification of the easier set (\textit{objects set}) using the qb SoftHand led to significantly worse results compared to the other grippers. We attribute this to the combination of the actuation mechanism and poor sensory feedback. This is an underactuated hand with only one motor controlling the action of the entire device by pulling a tendon running through all the fingers of the hand. Individual fingers thus passively conform to the shape of the object in the hand or close completely if they are free to move. Although this property is attractive for grasping, it is not effective for deformable object discrimination, since there is an ambiguous relationship between the feedback channels (motor position and current) and the shape and stiffness of the grasped object. On the other hand, 2-finger grippers can sense the dimension and stiffness along one dimension of the object. The Barrett Hand with three fingers is fully actuated and heavily sensorized providing the richest information about both shape and stiffness.

Works on robot haptic object classification typically employ a single device (robot hand or gripper), single sensory set, same grasping action, and one or few classifiers. This work is unique in that it tests four different devices with different action parameters and four different classifiers with different input types (with and without features) as well as the contribution of individual sensory modalities to the classification overall performance. While it was expected that the Barrett Hand as the most sophisticated and heavily sensorized device would perform best, we want to bring the following findings to the foreground. First, the choice of the classifier did not importantly change the results (see Fig.~\ref{fig:merged_classification}). SVM with hand-crafted features on input and LSTM on raw input time series performed best, although they use a very different approach. SVM was also much less computationally demanding. Second, grasping configuration significantly affected the classification performance on the qb SoftHand (\emph{a1} vs. \emph{a2} in Fig.~\ref{fig:merged_classification} (a)), while it had little effect on the Barrett Hand. There was a general trend across the devices that slower compression speeds improved classification. Generalization across grasping configurations and even speeds on the same device was limited (Fig.~\ref{fig:class_transfer}). Third, while combinations of sensors (position, effort, touch for the Barrett Hand) provide the best performance, their subsets often yield very good performance. Interestingly, gripper position only performed well in both parallel jaw grippers, as the evolution of object compression (strain) also reflects its stiffness, even if the effort (stress) is not measured (Fig.~\ref{fig:class_ablation}).

Direct quantitative comparison with related works is not possible, but we attempt a qualitative one. The work of Wang et al.~\cite{wang2022tactual} is related but complementary to ours. Using a 2-finger gripper with tactile sensors and force control grasping eight different deformable objects, they studied the effect of the force applied but randomized the pinching speed and hence could not study the effect thereof. On the stress-strain (force-indentation) curve, we compared hand-crafted features with raw time series, while \cite{wang2022tactual} presented features based on functional principal component analysis.
Liarokapis et al.~\cite{liarokapis2015unplanned} used a 3D printed object set to separate the shape and stiffness dimensions, resembling the object set used here. A two-finger hand with force sensors grasped the objects. They studied the effect of classifiers and features while we focused on gripper embodiment and actions.
Xu et al.~\cite{xu2013tactile} used tactile sensors to discriminate objects based on their material properties using the multimodal BioTac sensors. The superior performance of the Barrett Hand in our experiments and the results from using the touch sensors only that almost matched the full sensory set is intriguing, suggesting that a single ``haptic glance'' with capacitive pressure-sensitive sensors may suffice.

We complement the classification results with visualization of the sensory data collected while grasping the objects using PCA, providing a unique window into the different sources of variability. For every hand or gripper individually, the action parameters (hand configuration or compression speed) show up as the main source of variability, while the different objects grasped are responsible for the variance to a much smaller extent. This means that sensory data collected using grasping actions with different parameters should never be mixed together but either kept separately or the action parameters should be explicitly included in the classification process---similarly to the conclusions arrived at in \cite{Hoffmann_RAS_2014}. Furthermore, when data from multiple robot hands or grippers are combined, the clusters corresponding to the different devices are even more pronounced. This puts in question the possibility of learning or generalizing across robot embodiment (e.g., \cite{padalkar2023open}). Perhaps such ``knowledge transfer'' is possible on tasks where the embodiment is weak (e.g., visual inputs, robot commands in Cartesian space, task planning for only two-finger grippers), but there may be intrinsic limitations in haptic perception and grasping, where the sensory data is importantly shaped by the embodiment and the action parameters. 

\section{Future Work}
In this work we studied model-free single grasp deformable object discrimination exploiting mainly the mechanical response curve (stress-strain) obtained during a single object compression. Object stiffness can be sensed using simple parallel jaw grippers but also using tactile sensors (represented by the Barrett Hand in this work). The force-indentation curve could also be used to fit parametric models of elasticity and even viscoelasticity, which could be used for discrimination \cite{patni2024online}. On the other hand, there is a synergy between the recent advent of optical tactile sensors offering high-resolution tactile images (e.g., Gelsight~\cite{yuan2017gelsight}, Tactip~\cite{ward2018tactip}, Digit~\cite{lambeta2020digit}) and deep learning or transformer networks which can be applied to touch \cite{gao2024transformer}, but also to fuse data from different modalities, such as vision and touch \cite{han2021learning, li2023vito}. Hybrid approaches combining machine learning and parametric models may also be viable (e.g., \cite{spiers2016single}).

\bibliographystyle{IEEEtran}
\bibliography{objectClassificationSqueezing}

\vskip -2\baselineskip plus -1fil
% \begin{IEEEbiography}[{\includegraphics[width=1in,height=1.25in,clip,keepaspectratio]
\begin{IEEEbiography}[{\includegraphics[width=1in,height=1.25in,clip]{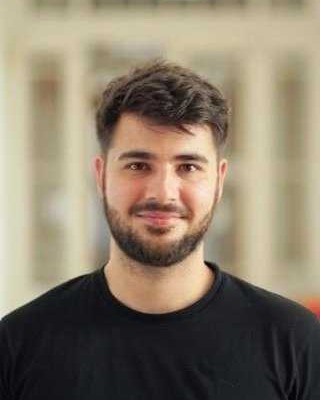}}]{Michal Pliska}
 is a PhD student in the Multi-robot Systems group (Department of Cybernetics, Faculty of Electrical Engineering, Czech Technical University in Prague) supervised by Assoc. Prof. Martin Saska. He received his MSc degree in Cybernetics and Robotics from the Czech Technical University in 2023. He is currently working toward the Ph.D. degree in Cybernetics and Robotics. His current research interests include multimodal state estimation and behavior prediction of UAVs, event-based vision, and planning through reinforcement learning.
\end{IEEEbiography}
\vskip -2\baselineskip plus -1fil
\begin{IEEEbiography}[{\includegraphics[width=1in,height=1.25in,clip]{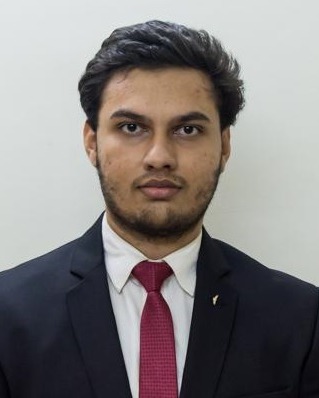}}]{Shubhan Patni}
 is a PhD student at the Humanoid and Cognitive Robotics Laboratory (Faculty of Electrical Engineering, CTU Prague) under Assoc. Prof. Matej Hoffmann. He graduated with a MSc. Robotics degree from the University of Bristol and the Bristol Robotics Laboratory (2019). He is currently working toward the Ph.D. degree in Informatics. At CTU Prague, he participated in the IPALM project (Interactive Perception-Action-Learning for Modelling Objects) and the ROBOPROX project (Robotics and advanced industrial production). His research interests are bio-inspired intelligence, embodied intelligence, and their application to robotics, especially tactile perception in humans and animals.
\end{IEEEbiography}
\vskip -2\baselineskip plus -1fil
\begin{IEEEbiography}[{\includegraphics[width=1in,height=1.25in,clip]{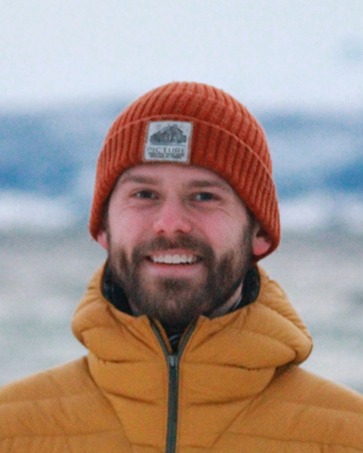}}]{Michal Mare\v{s}}
 holds a Master’s degree in Artificial Intelligence from FEE CTU (2023). His studies were concluded in 2023 by a thesis at Cisco Systems investigating the optimization of graph construction from network telemetry for graph machine learning algorithms. Currently, he is employed as a Data Scientist at Galytix, where he focuses on the explainability of anomalies and trends in financial data.
\end{IEEEbiography}
\vskip -2\baselineskip plus -1fil
\begin{IEEEbiography}[{\includegraphics[width=1in,height=1.25in,clip]{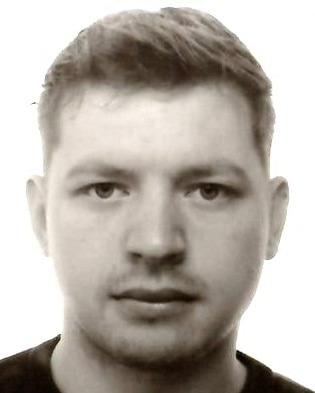}}]{Pavel Stoudek}
 received his MSc. degree in Cybernetics and Robotics in 2020 from FEE, Czech Technical University (CTU). He worked as a hardware engineer at the Multi-robot Systems group (Department of Cybernetics, Faculty of Electrical Engineering, CTU Prague). He is currently working as a robotics engineer in the Swarm team, Autonomous Robotics Research Center at Technology Innovation Institute (TII), Abu Dhabi. His interests are aerial robotics, humanoid robots, additive manufacturing, and JDM car culture.
\end{IEEEbiography}
\vskip -2\baselineskip plus -1fil
\begin{IEEEbiography}[{\includegraphics[width=1in,height=1.25in,clip]{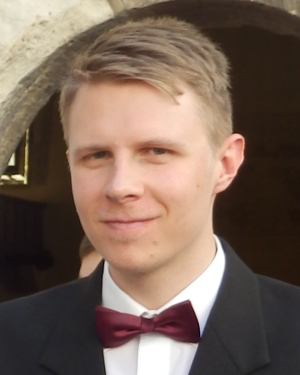}}]{Zdenek Straka}
received his Ph.D. in Artificial Intelligence and Biocybernetics from the Faculty of Electrical Engineering at the Czech Technical University in Prague (FEE CTU) in 2024. His current research interests include neural networks, neurorobotics, computational neuroscience, peripersonal space representations and applications of machine learning in healthcare. Dr. Straka received the ENNS Best Paper Award at ICANN 2017.
\end{IEEEbiography}
\vskip -2\baselineskip plus -1fil
\begin{IEEEbiography}[{\includegraphics[width=1in,height=1.25in,clip]{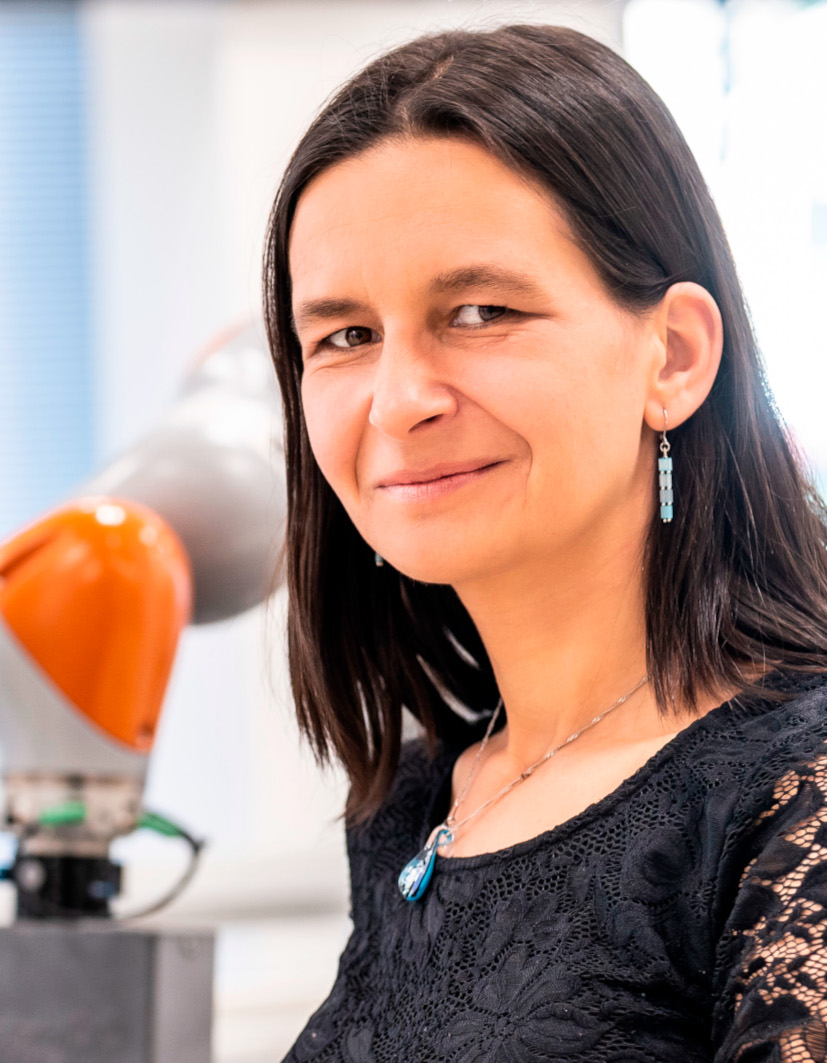}}]{Karla Stepanova} (Member, IEEE) received her Ph.D. in Artificial Intelligence and Biocybernetics from the Faculty of Electrical Engineering at the Czech Technical University in Prague (FEE CTU) in 2017. That same year, she joined the newly established Czech Institute of Informatics, Robotics, and Cybernetics at CTU in Prague (CIIRC CTU), where she is currently a researcher and co-founder of the Imitation Learning Centre. Since September 2024, she is also serving as a head of the Robotic Perception Group. Her research focuses on bio-inspired robotic perception, unsupervised learning, probabilistic multimodal models, and human-robot communication.
\end{IEEEbiography}
\vskip -2\baselineskip plus -1fil
\begin{IEEEbiography}[{\includegraphics[width=1in,height=1.25in,clip]{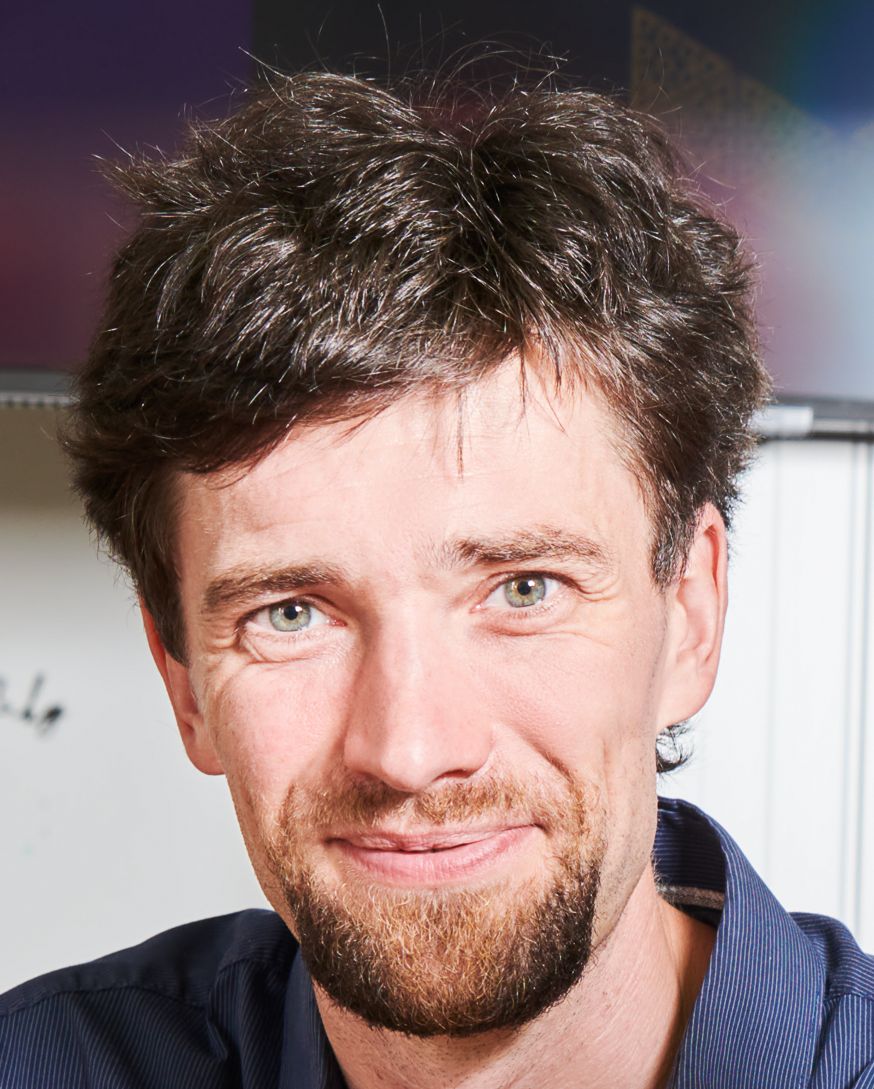}}]{Matej Hoffmann}
(Senior Member, IEEE) earned his PhD in Informatics at the Artificial Intelligence Lab, University of Zurich (2006-2012) and conducted postdoctoral research at the iCub Facility of the Italian Institute of Technology (2013-2016) supported by a Marie Curie Intra-European Fellowship. In 2017, he joined the Department of Cybernetics, Faculty of Electrical Engineering, Czech Technical University in Prague, where he is currently serving as Associate Professor and Coordinator of the Humanoid and Cognitive Robotics Group. His research interests include humanoid, cognitive developmental, and collaborative robotics as well as active perception for robot manipulation and grasping.
\end{IEEEbiography}
\vskip -2\baselineskip plus -1fil

\end{document}